\documentclass[lettersize,journal]{IEEEtran}
    
\usepackage{amsmath,amsfonts,amssymb}
\DeclareMathOperator*{\argmax}{arg\,max}
\usepackage{graphicx}
\usepackage{cite}
\usepackage{textcomp}
\usepackage{stfloats}
\usepackage{url}
\usepackage{verbatim}

\usepackage{algorithm}
\usepackage{algpseudocode}

\usepackage{array}
\usepackage{booktabs}
\usepackage{multirow}
\usepackage{tabularx}
\usepackage{siunitx}
\usepackage[export]{adjustbox}
\usepackage{pifont}
\usepackage{arydshln}
\usepackage{subcaption}
\usepackage{microtype}

\algrenewcommand\algorithmicrequire{\textbf{Require:}}
\algrenewcommand\algorithmicensure{\textbf{Ensure:}}
\newcolumntype{C}{>{\centering\arraybackslash}X}

\usepackage{hyperref} 

\begin{document}

\title{ODPure: Backdoor Purification for Object Detection via Ensemble Corruption Consensus}

\author{Li Zeng,
        Mingcheng Duan,
        Longfei Fan,
        Hangtao Zhang,
        Xianlong Wang,
        Yanchun Li,
        Xia Wen,
        Leo Yu Zhang, Member, IEEE

\thanks{Corresponding author: Yanchun Li (e-mail: ycli@xtu.edu.cn).}%
\thanks{L. Zeng is with the School of Computer and Communication Engineering, Changsha University of Science \& Technology, Changsha 410114, China.}%
\thanks{M. Duan, Y. Li, and X. Wen are with the School of Computer Science, Xiangtan University, Xiangtan 411105, China.}%
\thanks{L. Fan is with the School of Software, Yunnan University, Kunming 650500, China.}%
\thanks{H. Zhang is with the School of Cyber Science and Engineering, Huazhong University of Science and Technology, Wuhan 430074, China.}%
\thanks{X. Wang is with the Department of Computer Science, City University of Hong Kong, Hong Kong SAR, China.}%
\thanks{L. Y. Zhang is with the School of Information and Communication Technology, Griffith University, Southport, Queensland 4215, Australia.}%
}

\maketitle

\begin{abstract}
With the development of applications like autonomous driving, object detection has gained significant attention, while also highlighting critical vulnerabilities like backdoor attacks that severely compromise model integrity. Specifically, such attacks involve altering the categories of objects (\textit{i.e.,} object misclassification), removing bounding boxes (\textit{i.e.,} object disappearance), or generating bounding box proposals for non-existent objects (\textit{i.e.,} object generation) when a predefined trigger is present in the input. Although backdoor defenses for image classification are well-established, the research for object detection remains comparatively underexplored. Existing defenses address these threats by scanning outputs or models for potential backdoors but require discarding either malicious data or models. This remedy fails to enable a continuous and accurate perceptual stream for the object detection pipeline. To address such limitations, we propose ODPure, a novel input-stage black-box defense for object detection, which is based on input purification that ensures stable perception flows. Tailored to the dense prediction nature of object detectors, our \textit{Corruption-Reconstruction-Selection} (CRS) paradigm operates by neutralizing triggers through a diverse portfolio of corruptions to generate a massive pool of redundant proposals, then recovering fine-grained structural cues via generative priors, and finally employing voting to reach a consensus on the resulting detections. Comprehensive experiments demonstrate that our method provides robust defense against diverse backdoor attacks and trigger types while preserving baseline accuracy. Our code is available at \url{https://github.com/Alex66366/ODPure}.
\end{abstract}

\begin{IEEEkeywords}
Object detection, backdoor defense, input purification, robustness asymmetry, spatial consensus.
\end{IEEEkeywords}

\section{Introduction}\label{sec1}
\IEEEPARstart{R}{ecent} advancements in object detection, propelled by seminal works like Faster R-CNN~\cite{Ren2017Faster} and  YOLO~\cite{Redmon2016YOLO, Jocher2021YOLOv5}, have become crucial in safety-critical perception systems like autonomous driving~\cite{Chen2017MV3D, Li2022BEVFormer, Zou2023Survey}. However, object detection models are vulnerable to backdoor attacks~\cite{Chan2022Baddet, Cheng2024ODScan, Shen2023Django, Luo2023Untargeted, TrojAI-leaderboard}, causing them to exhibit malicious behaviors on trigger-containing samples while performing normally on clean samples. These malicious behaviors, including object misclassification, suppression of bounding boxes (\textit{i.e.,} object disappearance), and hallucinated detections of non-existent objects (\textit{i.e.,} object generation), compromise detectors.

To address these threats, existing defenses for object detection models typically intervene at two stages: \textit{post-inference output scanning}~\cite{Chan2022Baddet} and \textit{pre-deployment model inspection}~\cite{Cheng2024ODScan, Shen2023Django}. The former, exemplified by DetectorCleanse~\cite{Chan2022Baddet}, analyzes model outputs to identify and discard suspicious inputs. The latter, represented by ODSCAN~\cite{Cheng2024ODScan}, inspects the model's internal components to detect implanted backdoors before deployment. A shared limitation of both approaches is their exclusive focus on detecting threats~\cite{wang2026pvdetector}, leaving them with no recourse but to \textit{discard} the suspicious input or model. This leaves the crucial challenge of restoring or purifying poisoned samples for continued use unexplored, a gap our work aims to fill by proposing the first \textit{input-stage purification} defense for object detection.

\begin{figure}[t]
 \centering
 \subfloat[Clean scene]{ \includegraphics[width=0.18\textwidth]{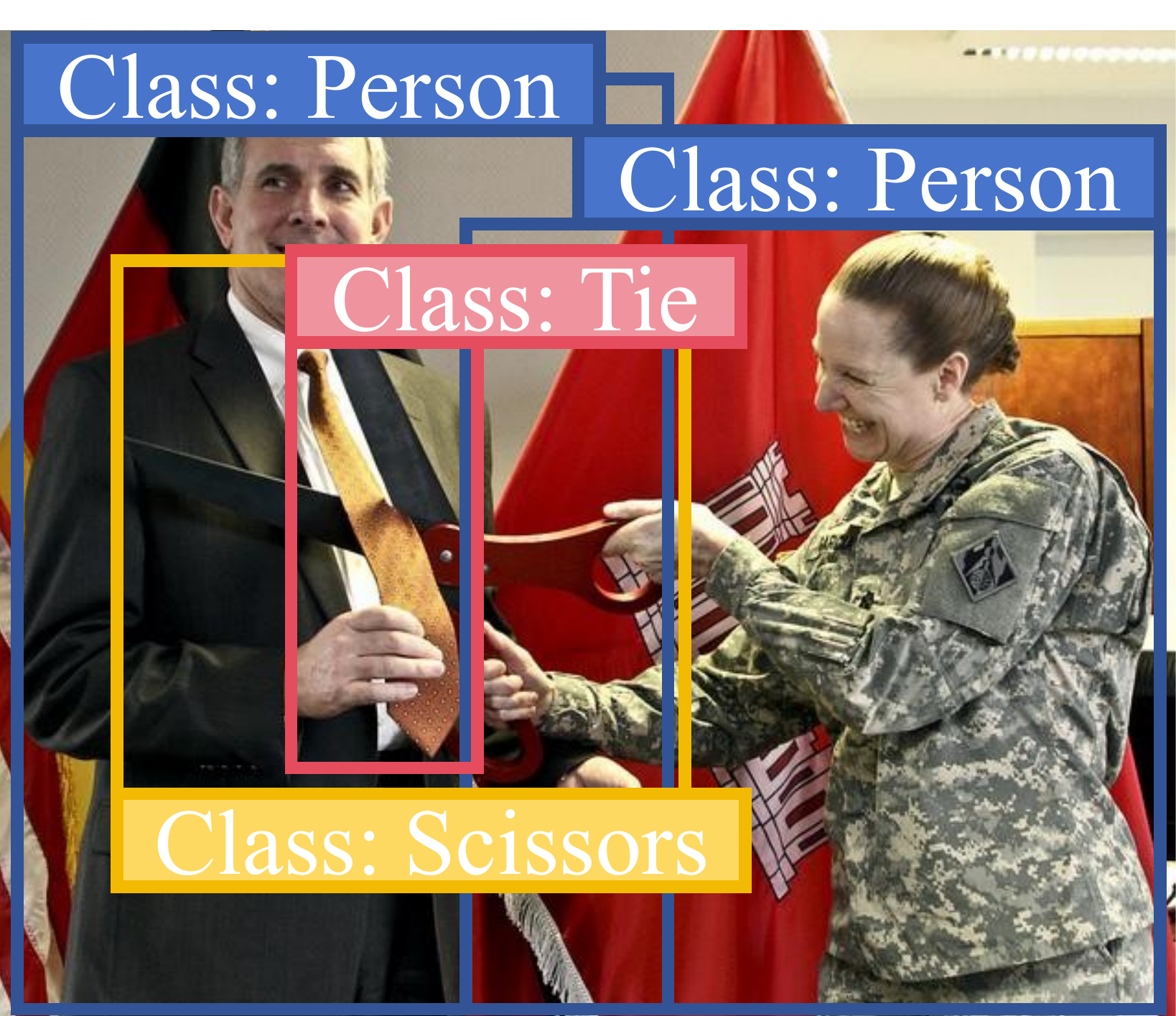} \label{fig1:clean}}
 \hspace{0.1in}
 \subfloat[Object Misclassification]{ \includegraphics[width=0.18\textwidth]{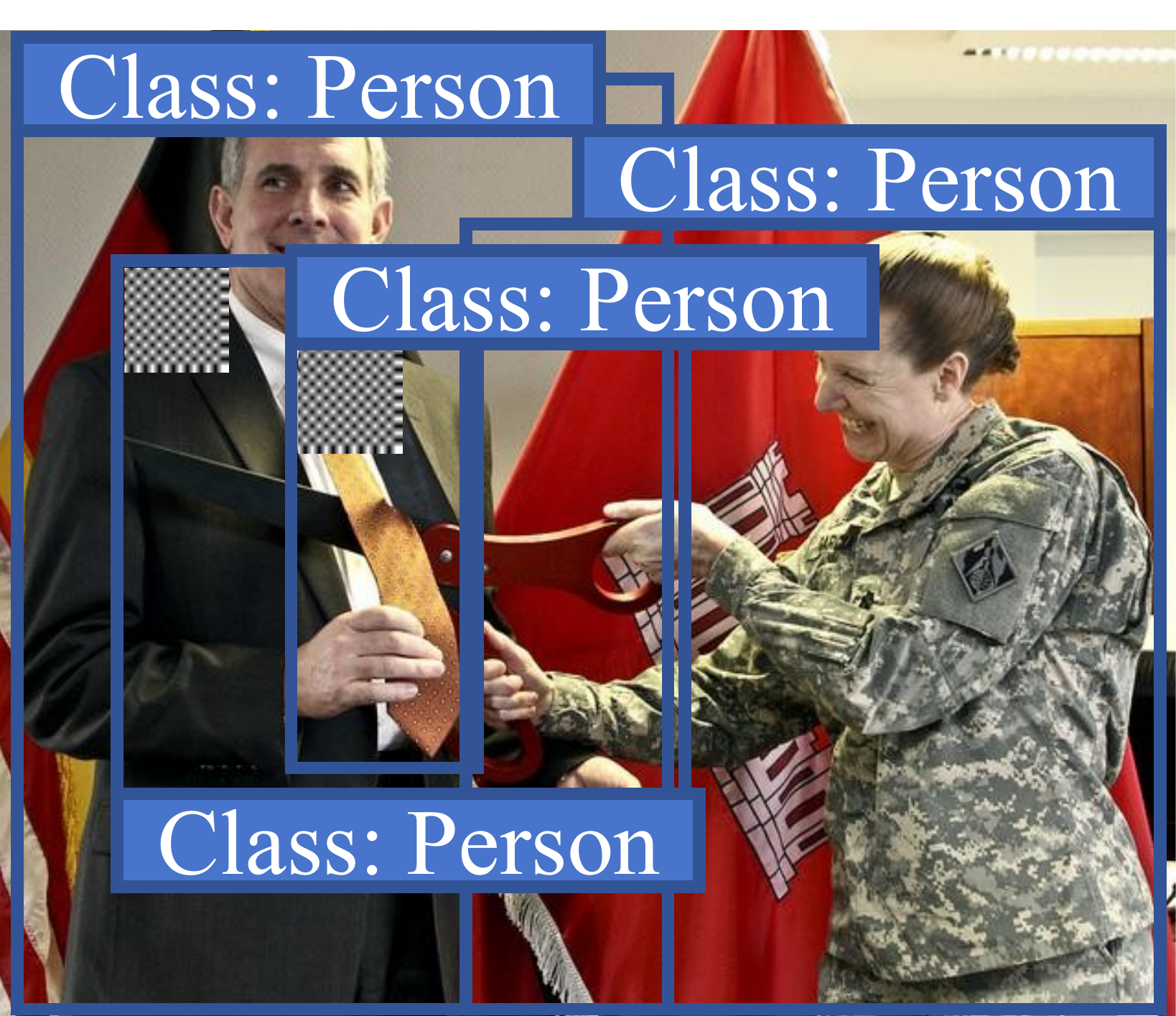} \label{fig1:misclassification}}
 \hspace{0.1in}
 \subfloat[Object Disappearance]{ \includegraphics[width=0.18\textwidth]{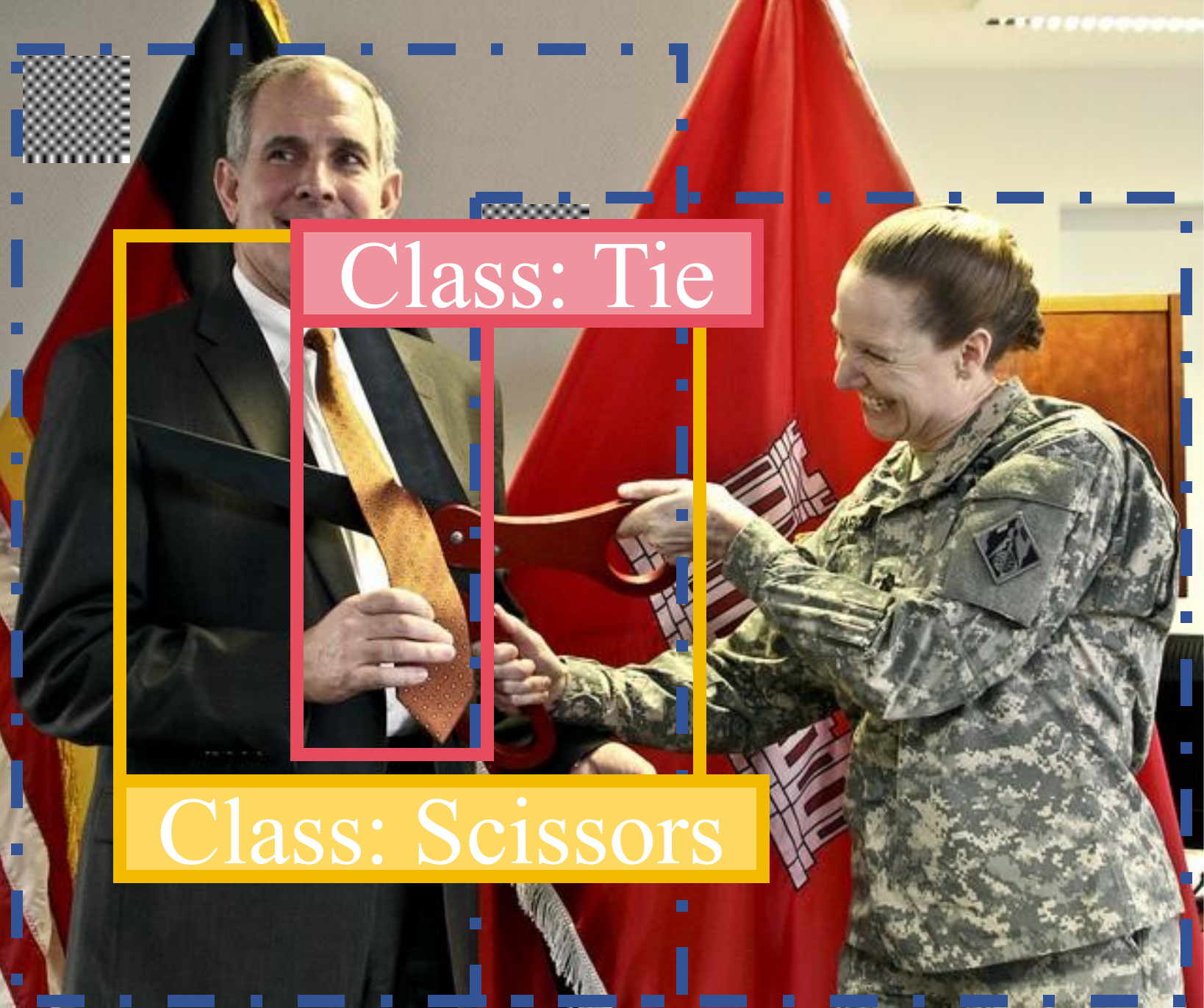} \label{fig1:disappearance}}
 \hspace{0.1in}
 \subfloat[Object Generation]{ \includegraphics[width=0.18\textwidth]{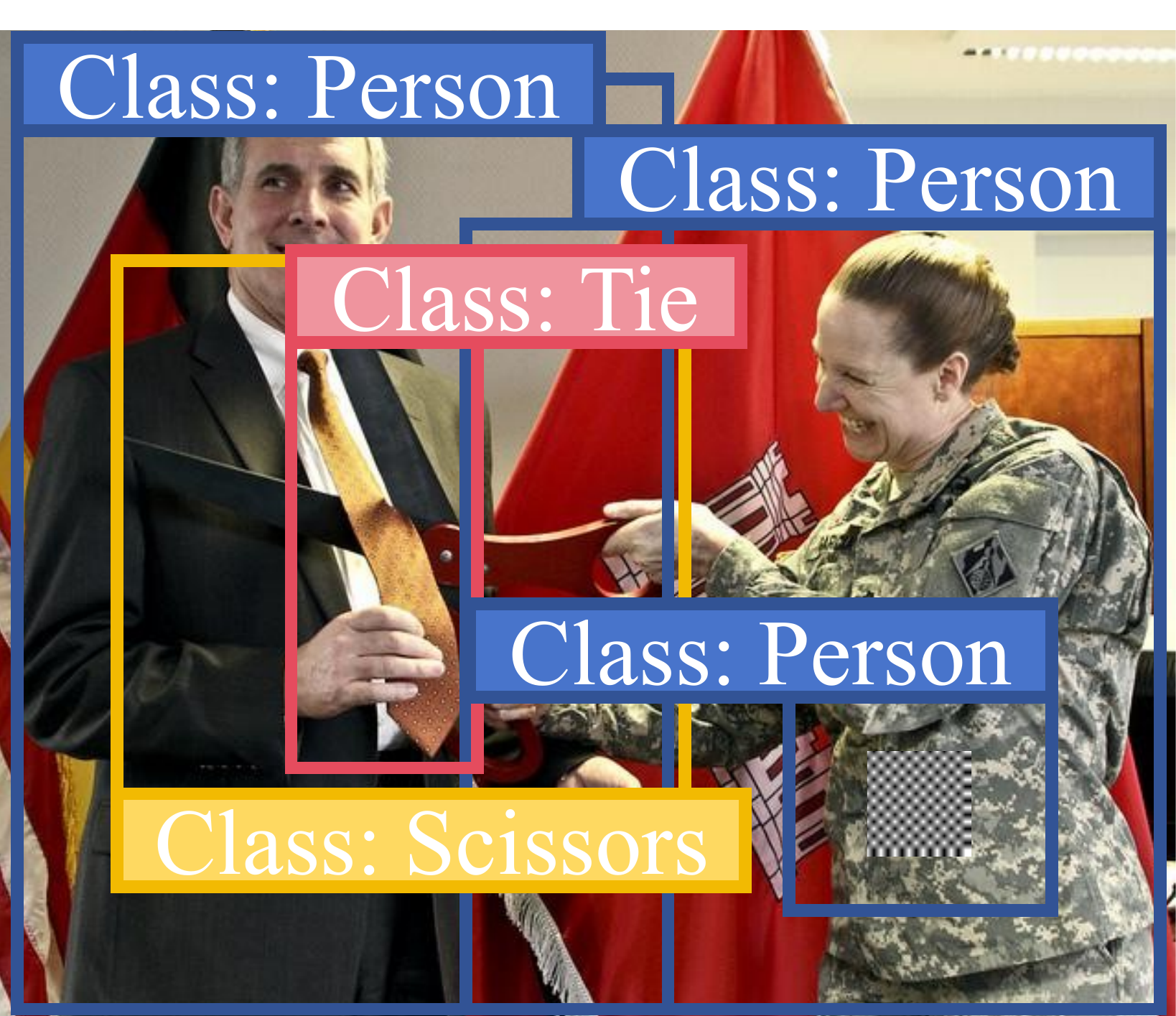} \label{fig1:generation}}
 \caption{Attack types (b)-(d) compared with clean scene (a).}
 \label{fig1:attack_types}
 \vspace{-2mm}
\end{figure}

While backdoor defenses have been explored in other fields like classification~\cite{Tarchoun2023Jedi, Wang2022Guided, Doan2020Februus, Yang2024SampDetox, Shi2023ZIP}, the peculiarities of attack behaviors in object detection (see Fig.~\ref{fig1:attack_types}) and their unique detection process present new challenges for backdoor defenses: (i) \textbf{Diversity of attack behaviors.} Unlike image classification, where backdoors typically cause a single type of failure (misclassification), attacks on object detectors can induce a more diverse spectrum of malicious behaviors. These include not only object misclassification but also object disappearance and object generation. This variety of potential attack outcomes makes defense significantly more challenging, as any effective solution must be robust enough to counter distinct failure modes. (ii) \textbf{Trade-off between attack disruption and multi-object preservation.} The defensive challenge in object detection is compounded by the sheer volume of proposals, which can be derived from tens of thousands of candidates (\textit{e.g.,} 20,000 anchors in RPN~\cite{Ren2017Faster}). This creates a difficult trade-off, where disrupting an attack on one object may inadvertently degrade the accuracy of benign detections, making it a significant hurdle to restore the detector's overall performance.

Our journey begins with two key observations. First, although these attack behaviors manifest as diverse failure patterns at different spatial locations and prediction heads, they are all driven by a single underlying cause, \textit{i.e.,} the backdoor trigger embedded in the image. In other words, we can mitigate multiple failure modes simply by neutralizing this trigger pattern. Second, we observe a key asymmetry in robustness consistency across the ensemble of corrupted variants. While benign object features degrade uniformly across different corruptions (\textit{i.e.,} \textit{consistent vulnerability}), backdoor triggers, functioning as artificial shortcuts, are hypersensitive to specific perturbation types (\textit{i.e.,} \textit{variable vulnerability}). This allows us to distill a consensus from a massive pool of redundant proposals. Specifically, benign objects consistently reappear and form stable clusters, whereas triggers exhibit unpredictable fluctuations, manifesting in some variants while vanishing in others. This instability prevents triggers from establishing a consensus, effectively filtering them out as noise.

Guided by these insights, we propose the \textit{Corruption-Reconstruction-Selection} (CRS) paradigm to address the identified challenges. To counter these diverse attack behaviors, the \textit{Corruption} stage targets their singular root cause by employing a rich portfolio of corruption types (\textit{e.g.,} Gaussian noise, defocus blur, snow) at multiple severities to ensure trigger neutralization. To resolve the security-utility trade-off, we leverage the observed robustness asymmetry: the \textit{Reconstruction} stage utilizes a diffusion-based restoration model~\cite{Lin2024DiffBIR,wangsurvey,zeng2025psfd,xie2026typo} to recover consistently degraded fine-grained details via generative priors, while the \textit{Selection} stage employs a consensus-based selection mechanism (implemented via density-based clustering and confidence-based voting) to exploit spatial prediction consistency, distilling robust detections and filtering out unstable trigger-induced artifacts. We operationalize this paradigm in ODPure, a novel input-stage, black-box defense. Crucially, instead of the detect-and-discard strategy used by output-scanning methods~\cite{Chan2022Baddet}, our CRS paradigm adopts a purification-based defense tailored to dense, multi-instance object detection. It sanitizes inputs while preserving scene semantics and instance locality, so that the detector can still produce usable proposals, thereby maintaining downstream decision-making in safety-critical perception pipelines (\textit{e.g.,} autonomous driving and cloud-assisted vehicular audits). Empirically, ODPure demonstrates exceptional efficacy against severe threats. For instance, under the devastating Object Misclassification Attack on the Microsoft COCO (COCO)~\cite{Lin2014COCO} dataset where the model utility collapses to 0.4\%, ODPure successfully restores the mean Average Precision (mAP) to 52.0\% and suppresses the Attack Success Rate (ASR) to 1.5\%. This performance significantly outperforms state-of-the-art classification-based baselines such as ZIP~\cite{Shi2023ZIP}. Overall, the main contributions of this paper are as follows:
\begin{itemize}
    \item We introduce the CRS paradigm, the first input-stage purification framework for object detection. This paradigm resolves the inherent trade-off between attack disruption and multi-object preservation, enabling continuous and accurate perceptual streams.
    \item We propose ODPure, a black-box defense that operationalizes the CRS framework. By integrating diverse corruptions, generative restoration, and consensus-based selection, ODPure functions in a fully model-agnostic manner. This nature makes it highly practical for scenarios like \textit{Machine-Learning-as-a-Service} (MLaaS)~\cite{Ribeiro2015Mlaas}.
    \item Extensive experiments demonstrate that ODPure breaks the trade-off limiting classification-based purification methods. By decoupling trigger neutralization from multi-object preservation, our method achieves state-of-the-art performance without compromising detection utility.
\end{itemize}
\section{Related Work}\label{sec2}
\subsection{Attacks on Object Detectors}
Unlike standard image classification, object detection performs a dual task: simultaneously identifying an object's class and determining its precise location via regression~\cite{Ren2017Faster, Redmon2016YOLO, Zou2023Survey}. This dual objective inherently expands the attack surface, enabling more complex and malicious backdoor behaviors~\cite{Chan2022Baddet, Cheng2024ODScan, Shen2023Django, Luo2023Untargeted, TrojAI-leaderboard}. Specifically, beyond standard misclassification errors, backdoored detectors can exhibit sophisticated, task-specific failures, such as \textit{object disappearance} (failing to detect existing targets) or \textit{object generation} (hallucinating false objects in empty spaces), as illustrated in Fig.~\ref{fig1:attack_types}. We describe these attack behaviors below.

\noindent \textbf{Object Misclassification Attack (OMA).~\cite{Chan2022Baddet, Cheng2024ODScan, Shen2023Django}}
Triggered by $t$, the model $M$ is forced to assign an incorrect label $\hat{l}_i$ to a targeted object while maintaining the accurate bounding box $\hat{b}_i$ (see Fig.~\ref{fig1:misclassification}).

\noindent \textbf{Object Disappearance Attack (ODA).~\cite{Chan2022Baddet, Cheng2024ODScan, Luo2023Untargeted, Shen2023Django}}
Triggered by $t$, the model $M$ suppresses the detection of a targeted object, failing to generate the prediction $(\hat{b}_i, \hat{l}_i)$ (see Fig.~\ref{fig1:disappearance}).

\noindent \textbf{Object Generation Attack (OGA).~\cite{Chan2022Baddet, Cheng2024ODScan, Shen2023Django}}
Triggered by $t$, the model $M$ generates a false-positive ``ghost'' detection $(\hat{b}_f, \hat{l}_f)$ in a specified region (see Fig.~\ref{fig1:generation}).
\subsection{Backdoor Defenses for Object Detectors}
Unlike the extensive body of work on backdoor defenses for classification tasks, research on defending object detectors remains comparatively underdeveloped. Model-scanning approaches, exemplified by ODSCAN~\cite{Cheng2024ODScan}, inspect the model's internal parameters to detect implanted backdoors. However, their reliance on a white-box setting renders them impractical in many real-world scenarios, such as MLaaS~\cite{Ribeiro2015Mlaas}, where users lack access to model internals. In contrast, output-scanning defenses such as DetectorCleanse~\cite{Chan2022Baddet} analyze model predictions to identify and discard potentially malicious inputs. This approach reveals a prevailing focus in the literature on \textit{backdoor detection}, while the specific challenge of \textit{backdoor purification} for object detection remains a notable research gap. To bridge this gap, we introduce an input-stage purification method. Instead of discarding the input, our method purifies the sample, ensuring that the sanitized data remains available for downstream decision-making. Doing so preserves the operational integrity of critical deep perception pipelines.

\subsection{Input-Stage Purification in Classification}
Input-stage purification is an active research frontier, widely applied in defenses against both adversarial attacks and backdoor attacks~\cite{Tarchoun2023Jedi, Wang2022Guided, Doan2020Februus, Yang2024SampDetox, Shi2023ZIP,Wang2024ECLIPSE,zhang2026defending,zhou2025darkhash,wang2026advedm,wang2025breakingphysical,song2025pb,yao2024reverse}. Broadly, adversarial purification~\cite{Tarchoun2023Jedi, Wang2022Guided,zeng2026ghostprompt} aims to remove imperceptible perturbations from inputs at inference time, utilizing strategies ranging from the statistical analysis of high-entropy regions~\cite{Tarchoun2023Jedi} to generative restoration~\cite{Wang2022Guided}. In contrast, backdoor purification, the focus of this work, seeks to neutralize embedded triggers. Research in this area has generally followed two paradigms. Early approaches focused on localizing and repairing trigger regions, often using visual explanations like Grad-CAM~\cite{Selvaraju2017GradCAM} to find the trigger and generative inpainting to remove it~\cite{Doan2020Februus}. However, these methods are often tailored to explicit, localized triggers (\textit{e.g.,} patches) and risk failing against more subtle, global triggers. To address these limitations and handle a wider range of attacks, attention has shifted toward global generative restoration via diffusion models~\cite{Yang2024SampDetox, Shi2023ZIP}.

While highly effective for image classification, these generative approaches face a critical, unresolved conflict between their classification-first design philosophy and the multi-object nature of detection. Unlike classification (which makes a single prediction), a detector must simultaneously preserve numerous, diverse benign instances. Current generative methods, however, typically rely on fixed purification strength configurations (\textit{e.g.,} a specific noise level or blur kernel), prioritizing the disruption of the trigger's influence on the final class label at the expense of meticulous image reconstruction. For instance, while ZIP~\cite{Shi2023ZIP} explored using multiple transformation types (\textit{e.g.,} blur and grayscale), it restricted them to a fixed intensity (\textit{e.g.,} a specific kernel size), neglecting the potential of varying severities. This simplistic approach still creates an impossible dilemma: the purification strength is inevitably either (i) \textit{too weak} to neutralize the backdoor, or (ii) \textit{too intense}, destroying the legitimate, high-frequency details (\textit{e.g.,} sharp edges, textures) essential for precise localization.
This fundamental trade-off between attack disruption and multi-object preservation presents a significant challenge for high-stakes perception tasks.

\section{Preliminaries}\label{sec3}
\subsection{Threat Model}

Following established threat models for backdoor attacks~\cite{Chan2022Baddet, Cheng2024ODScan,zhang2024detector,wang2024trojanrobot}, we assume the primary backdoor attack vector is data poisoning~\cite{Gu2017BadNets, Chen2017Targeted, Chan2022Baddet,li2025fine,zhang2023denial}, where the attacker's objective is to contaminate the dataset to implant a hidden backdoor into the final detector.

\noindent \textbf{Defender's Constraints.}
To reflect realistic security scenarios, we assume the defender operates under a strict set of constraints, distinguishing our work from white-box approaches:
\begin{itemize}
    \item \textbf{Attack-Agnostic:} The defender has zero prior knowledge of the attack configuration. They are unaware of the trigger's existence, its attributes (\textit{e.g.,} pattern, size, location), or the malicious behavior it aims to induce.
    \item \textbf{Strict Black-Box Access:} The defender is limited to query-only access to the model, simulating real-world MLaaS scenarios~\cite{Ribeiro2015Mlaas}. They can only feed input samples to the model and observe the output (\textit{i.e.,} bounding boxes and class scores). Critical internal information (\textit{e.g.,} architecture, parameters, gradients) is strictly inaccessible.
    \item \textbf{Data-Agnostic:} The defender has no access to the original training dataset (neither clean nor poisoned) and possesses no isolated trigger samples for analysis. The defense must operate effectively at inference time solely based on the current input.
\end{itemize}

\noindent \textbf{Defender's Goals.}
Given these constraints, an effective defense must satisfy two primary objectives:
\begin{itemize}
    \item \textbf{Security:} The defense must effectively neutralize the backdoor, preventing malicious behaviors on triggered inputs. This is quantified by achieving a low ASR.
    \item \textbf{Fidelity:} The defense must maintain high utility for downstream tasks. This is measured by the mAP on two fronts: (i) preserving the accuracy of benign inputs and (ii) restoring the detectability of purified malicious inputs.
\end{itemize}

\subsection{Detection Task and Performance Metrics}\label{sec3-2}
An object detection model $M$ identifies objects (location, size, and class) within an input image $I$. Given an input image $I$ with $p$ objects, the ground-truth annotations are defined as $\{(b_i, l_i)\}_{i=1}^p$, where $b_i$ represents the bounding box of the $i$-th object. Specifically, $b_i$ is a rectangular bounding box described by the coordinates of its top-left and bottom-right corners: $b_i = (b_i^{x_1}, b_i^{y_1}, b_i^{x_2}, b_i^{y_2})$. The variable $l_i$ denotes the ground-truth class of the $i$-th object.

For a given input $I$, the object detection model $M$ predicts a set of detections: $M(I) = \{\hat{r}_j\}_{j=1}^N = \{(\hat{b}_j, \hat{l}_j, \hat{s}_j)\}_{j=1}^N$. Here, $N$ is the total number of detections, $\hat{b}_j$ is the predicted bounding box, $\hat{l}_j$ is its predicted class, and $\hat{s}_j$ is the confidence score. The objective of training is to generate detections that accurately match the ground-truth annotations. This match is defined by the Intersection over Union (IoU)~\cite{Everingham2010VOC} exceeding a threshold $\tau$. An ideal model would find a detection $\hat{r}_j$ for every ground-truth object $b_i$ such that:
\begin{equation}
    \text{IoU}(\hat{b}_j, b_i) \ge \tau \quad \text{and} \quad \hat{l}_j = l_i.
\end{equation}

\begin{figure}[t]
    \centering
    \includegraphics[width=0.90\columnwidth]{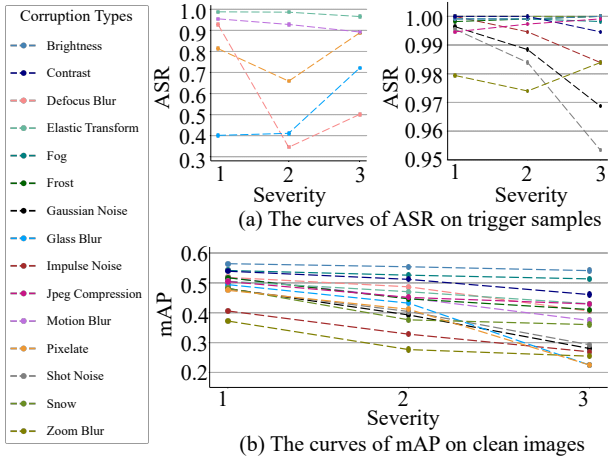}
    \caption{ASR and mAP vs. corruptions for an ODA-attacked YOLO model~\cite{Redmon2016YOLO, Jocher2021YOLOv5} under a standard $29\times29$ chessboard trigger, evaluated on (a) the poisoned COCO dataset (ASR) and (b) the clean COCO dataset (mAP).}
    \label{fig:asr_corruption_impact}
\end{figure}

\noindent \textbf{Detector Performance.} Typically, model utility is measured by mAP~\cite{Zou2023Survey}. In this work, we specifically adopt the mAP at an Intersection over Union (IoU) threshold of 0.5 (\textit{i.e.,} mAP@0.5) to evaluate detection accuracy.

\noindent \textbf{Attack Effectiveness.} This is gauged by the ASR, following the definition in~\cite{Chan2022Baddet}. This metric measures the proportion of successful attacks out of the total number of trigger-backdoor activation attempts.

\section{Motivation}\label{4}

\begin{figure}[t] 
\centering
\includegraphics[width=0.7\columnwidth]{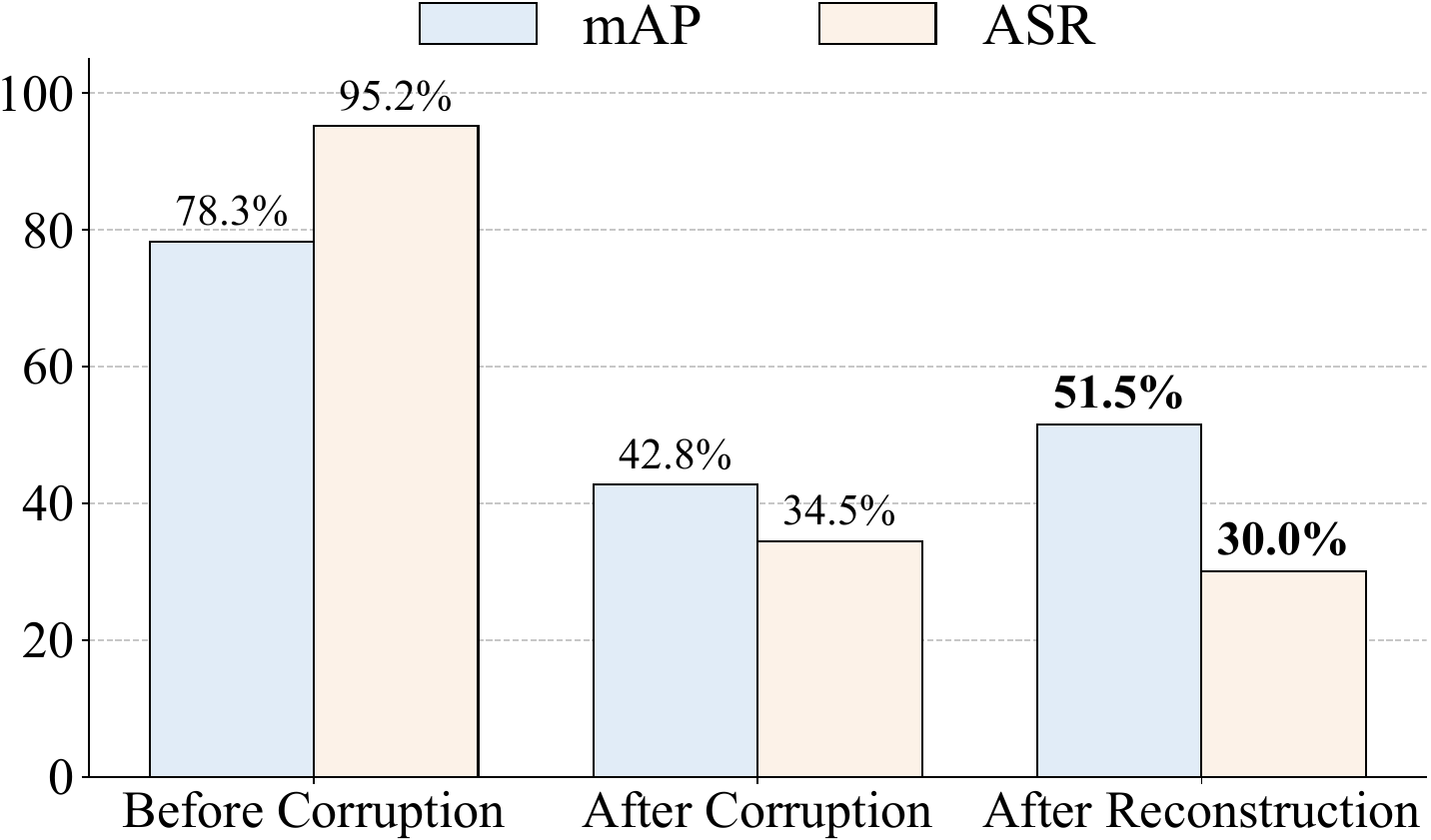} 
\caption{Quantitative validation of \textit{Reconstruction}, showing the stage-wise performance. The experiment uses the same ODA-attacked model and poisoned COCO dataset from Fig.~\ref{fig:asr_corruption_impact}, illustrating the specific defocus blur (severity 2) case.}
\label{fig:recon_motivation}
\vspace{-2mm}
\end{figure}

\subsection{Motivation for Purification}\label{4-1}

To address the diversity of attack behaviors identified in Section~\ref{sec1}, we focus on the singular root cause: the backdoor trigger. Disrupting these trigger characteristics effectively neutralizes all attack types (OMA, ODA, OGA) simultaneously. However, directly applying existing purification methods in classification~\cite{Yang2024SampDetox, Shi2023ZIP} fails to address the trade-off between attack disruption and multi-object preservation. Such methods face an inescapable dilemma where the intervention is either (i) \textit{too weak} to neutralize the trigger, or (ii) \textit{too intense}, destroying fine-grained benign features. Fortunately, the empirical results in Fig.~\ref{fig:asr_corruption_impact} reveal a critical \textit{robustness asymmetry} that allows us to decouple these conflicting objectives. Fig.~\ref{fig:asr_corruption_impact}(a) (separated into high- and low-impact groups for clarity) demonstrates the highly \textit{variable vulnerability} of backdoor triggers. Specifically, while certain corruptions (\textit{e.g.,} defocus blur in the left plot) drastically disrupt the trigger mechanisms, causing a sharp drop in ASR, others (\textit{e.g.,} contrast in the right plot) fail to neutralize the trigger, leaving the ASR nearly unchanged. In stark contrast, Fig.~\ref{fig:asr_corruption_impact}(b) reveals a \textit{consistent vulnerability} of benign objects, as the mAP exhibits a uniform degradation pattern across different corruptions. This inconsistency, also observed in classification~\cite{Liu2023Detecting}, arises because the trigger acts as a simple, repetitive model \textit{shortcut} that deviates from natural features~\cite{Yang2024SampDetox}, reacting erratically to different noises. Conversely, the complex, diverse features of clean objects degrade uniformly.

Driven by this evidence, we propose a fundamental shift from detection to purification. Unlike prior works~\cite{Liu2023Detecting} that utilize this asymmetry to identify poisoned samples in image classification, we leverage it to establish our CRS paradigm for restoring them in the more challenging context of object detection. First, to exploit the trigger's \textit{variable vulnerability}, we employ a diverse portfolio of \textit{Corruptions} to ensure neutralization. Second, leveraging the observed asymmetry, we utilize high-fidelity \textit{Reconstruction} to selectively eliminate the trigger, treating it as \textit{out-of-distribution} noise, while recovering the consistently degraded clean features (validated in Section~\ref{sec4-2}). Finally, a robust \textit{Selection} mechanism aggregates these predictions to distill the final output.

\subsection{Motivation for High-Fidelity Reconstruction}\label{sec4-2}

While the \textit{Corruption} phase effectively disrupts the trigger's influence, it inevitably introduces severe degradation to benign object features, rendering the detector impractical. To address this, we leverage the distinct manifold properties of benign objects versus backdoor triggers. Benign objects possess robust semantic structures that align with the generative priors of diffusion models~\cite{Lin2024DiffBIR}, exhibiting restorable consistency even after severe corruption. In contrast, backdoor triggers appear as high-frequency, \textit{out-of-distribution} artifacts. This asymmetry creates a filtering opportunity: diverse corruptions disrupt the trigger's fragile patterns (\textit{variable vulnerability}), preventing the restoration model from consistently recovering them (\textit{i.e.,} suppressing it during the denoising process). Consequently, while clean objects are faithfully reconstructed across all variants, triggers fail to survive consistently, allowing them to be easily discarded during the final proposal selection.

\begin{figure}[t]
\centering
\subfloat[Clean scene]{%
    \includegraphics[width=0.46\columnwidth]{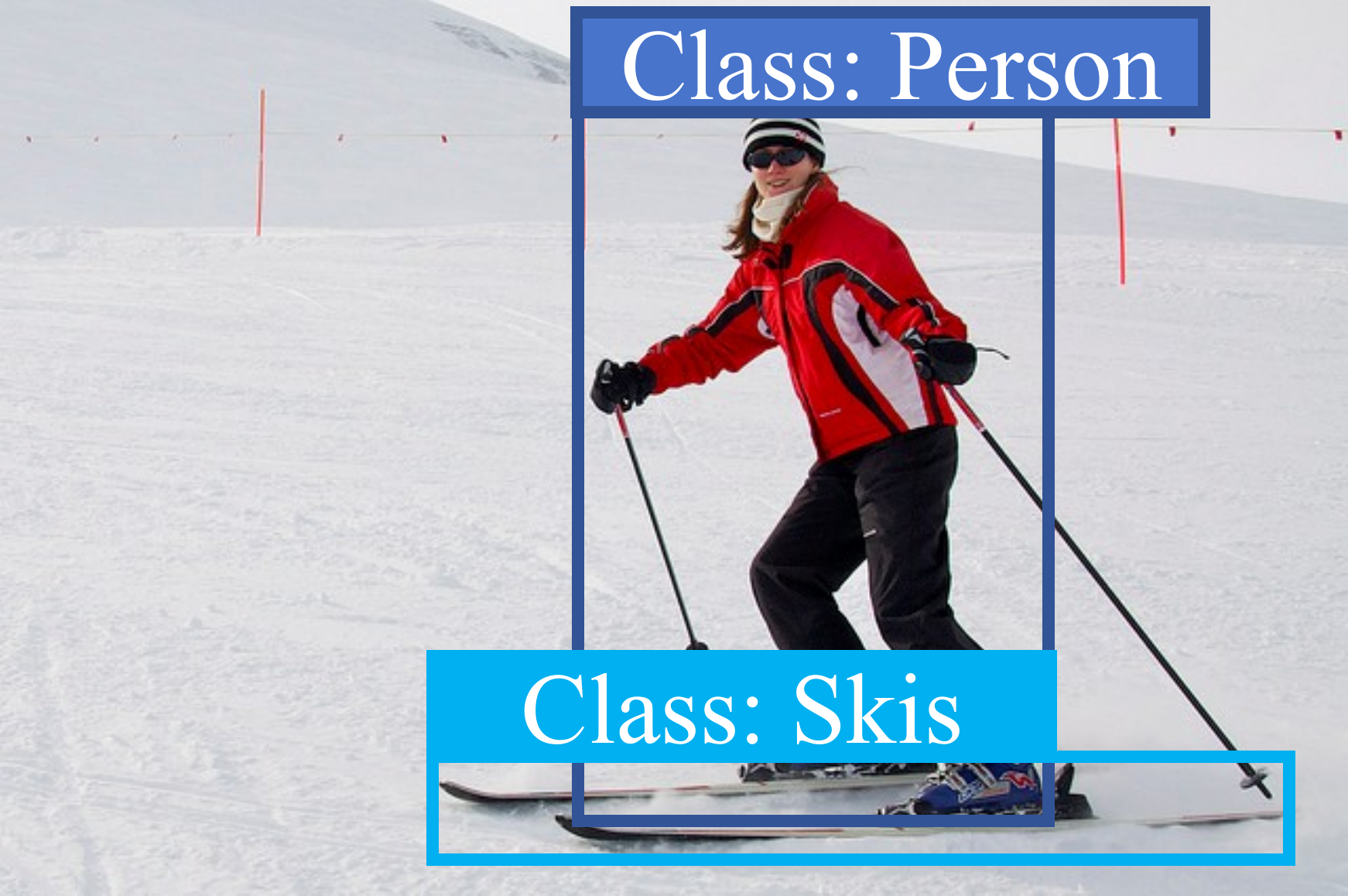}%
    \label{fig:clean}%
}
\hfill
\subfloat[Poisoned scene]{%
    \includegraphics[width=0.46\columnwidth]{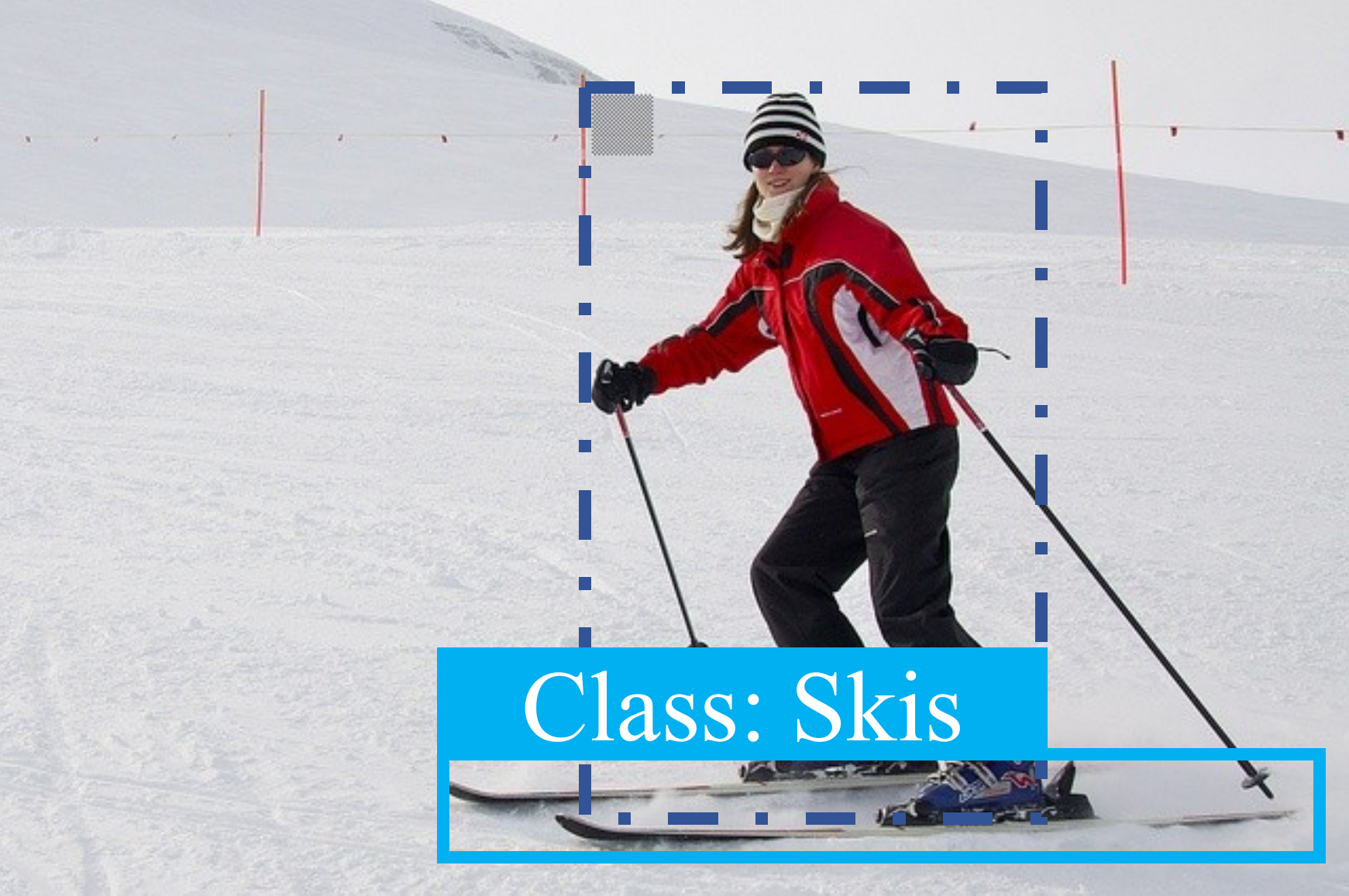}%
    \label{fig:misclassification}%
}
\\ \vspace{1mm}
\subfloat[After corruption]{%
    \includegraphics[width=0.46\columnwidth]{reconstruction_3.pdf}%
    \label{fig:disappearance}%
}
\hfill
\subfloat[After reconstruction]{%
    \includegraphics[width=0.46\columnwidth]{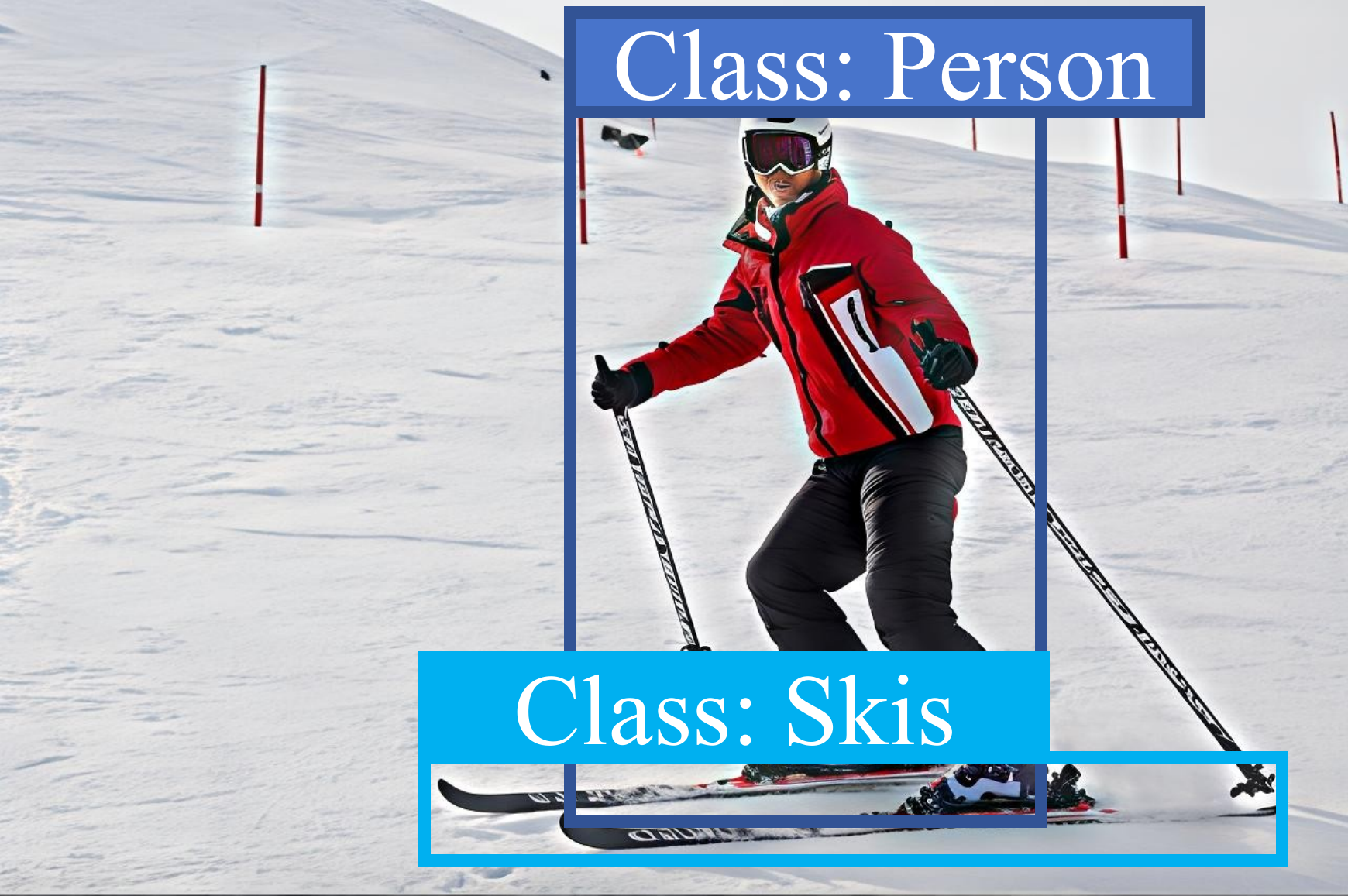}%
    \label{fig:generation}%
}
\vspace{-1mm}
\caption{Visual validation of \textit{Reconstruction}, demonstrated with an ODA example. The experimental settings are identical to those in Fig.~\ref{fig:recon_motivation}.}
\label{fig:reconstruction_motivation}
\vspace{-2mm}
\end{figure}

Empirical results (Fig.~\ref{fig:recon_motivation}) also validate this hypothesis: applying only the \textit{Corruption} step causes a catastrophic mAP drop (78.3\% to 42.8\%), whereas the subsequent \textit{Reconstruction} phase effectively restores the mAP to 51.5\% while further suppressing the ASR (34.5\% to 30.0\%). The visual evidence in Fig.~\ref{fig:reconstruction_motivation} confirms this dual capability: trigger artifacts are eliminated while fine-grained object details are faithfully reconstructed for detection.

\begin{figure*}[ht]
    \centering
    \includegraphics[width=\textwidth]{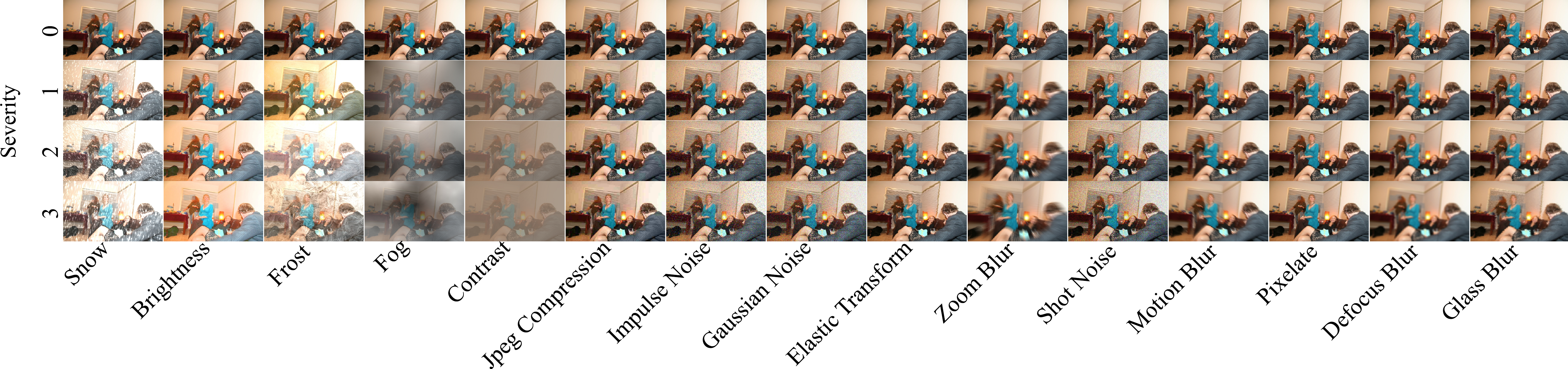} 
    \caption{Examples of image corruptions at severity levels 0 (original) to 3 on PASCAL VOC (VOC)~\cite{Everingham2010VOC}.}
    \label{fig:corruption_effect}
    \vspace{-2mm}
\end{figure*}

\section{Methodology}
\label{5}

\begin{figure*}[!t]
    \centering
    \includegraphics[width=0.78\textwidth]{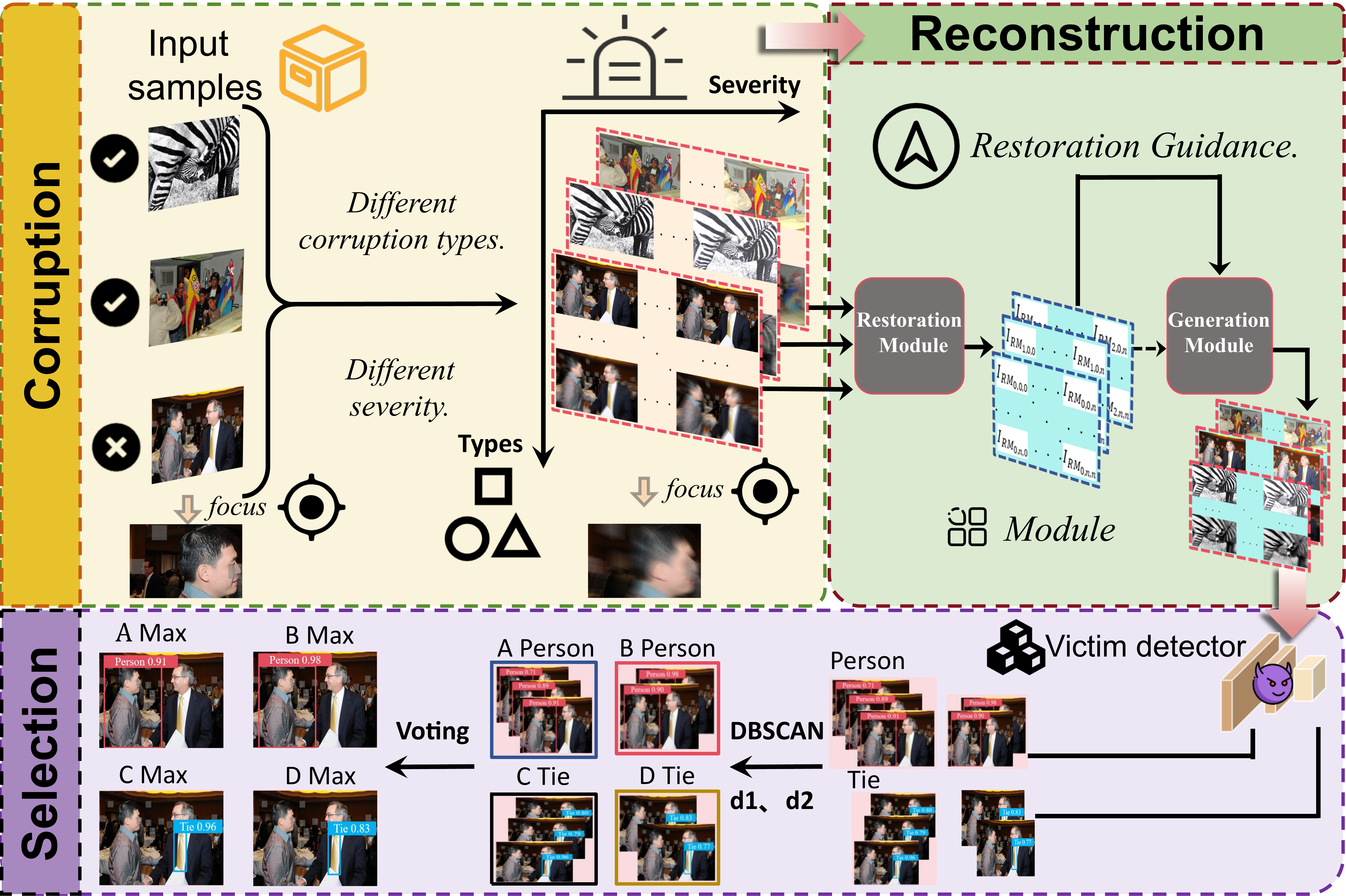}
    \caption{Overview of the ODPure pipeline. It operationalizes the CRS paradigm by cascading (i) \textit{Corruption} for trigger disruption, (ii) \textit{Reconstruction} for feature restoration, and (iii) \textit{Selection} for consensus distillation.}
    \label{fig:pipeline}
    \vspace{-2mm}
\end{figure*}

As illustrated in Fig.~\ref{fig:pipeline}, ODPure operationalizes the CRS paradigm through three sequential stages:
(i) \textit{Corruption} (Section~\ref{5-1}), which applies a diverse portfolio of corruption functions to disrupt the structural integrity of backdoor triggers by exploiting their variable vulnerability;
(ii) \textit{Reconstruction} (Section~\ref{5-2}), which leverages generative priors to repair regression-critical features degraded during corruption, ensuring the input is suitable for dense prediction; and
(iii) \textit{Selection} (Section~\ref{sec:selection_methodology}), which aggregates the massive pool of proposals from the purified ensemble and employs a consensus-based mechanism to filter out unstable trigger anomalies while distilling high-confidence benign detections.

\subsection{Corruption: Disrupting Trigger Characteristics}
\label{5-1}

To disrupt trigger characteristics for defensive purposes, our ODPure leverages image corruption. However, as demonstrated in our motivational analysis in Section~\ref{4-1}, relying on limited corruption types or fixed severity configurations is an unreliable strategy due to their trigger-dependent effectiveness.

Therefore, to ensure robustness, our approach is built upon applying a diverse portfolio of corruptions. We formally define this process as the application of a set of corruption functions, $\mathcal{F}$, to a potentially malicious input image $I$. Each function $f_{i,j} \in \mathcal{F}$ represents the $i$-th corruption type applied at the $j$-th severity level. This transforms the single input $I$ into a set of corrupted variants, $\mathcal{I}_{corr}$, defined as:
\begin{IEEEeqnarray}{rCl}
    \mathcal{I}_{corr} & = & \{ f_{i,j}(I) \mid 1 \le i \le n, \, 1 \le j \le m \}.
    \label{eq:corruption_process}
\end{IEEEeqnarray}
The application of this corruption process yields a diverse set of image variants, the visual effects of which are illustrated in Fig.~\ref{fig:corruption_effect}. The effectiveness of this multi-faceted corruption strategy, both as a standalone defensive step and as a critical component of our ODPure, is quantitatively analyzed in our ablation studies in Section~\ref{sec:ablation}.

\subsection{Reconstruction: Restoring Regression-Critical Features}
\label{5-2}
While the \textit{Corruption} phase effectively disrupts the trigger patterns, it inevitably introduces noise that degrades the high-frequency spatial information, such as edges and textures, essential for object detection. Unlike image classification, object detection is a dual-task problem requiring both accurate classification and precise bounding box regression~\cite{Ren2017Faster,zhang2025test}. The degradation caused by corruption severely impairs the detector's regression branch, leading to drifted anchors and reduced IoU with ground-truth objects.

To recover these regression-critical features, we employ an off-the-shelf, pre-trained image restoration model adapted from DiffBIR~\cite{Lin2024DiffBIR}. Crucially, this model is pre-trained on large-scale external datasets (\textit{e.g.,} ImageNet~\cite{Deng2009ImageNet}) to learn universal image priors, allowing it to function as a general-purpose denoiser on the target domain (\textit{e.g.,} COCO and VOC) in a zero-shot manner, fully adhering to our data-agnostic threat model.

The restoration process operates in two sequential inference stages, leveraging the objectives optimized during the model's pre-training.

First, a Restoration Module (RM) is utilized to remove content-independent degradations. This module was pre-trained to minimize the Mean Squared Error (MSE):
\begin{equation}
    \mathcal{L}_{RM} = \|I_{RM} - I_{gt}\|_{2}^{2},
\end{equation}
where $I_{gt}$ denotes the ground-truth images from the external pre-training dataset. In our inference pipeline, this stage recovers the coarse structural layout essential for subsequent anchor alignment.

Second, to prevent the over-smoothing issue common in standard denoising~\cite{Ledig2017SRGAN}, we utilize a Generation Module (GM) that leverages the frozen generative prior of Stable Diffusion~\cite{Rombach2022High}. The GM was pre-optimized via the latent diffusion objective:
\begin{equation}
    \mathcal{L}_{GM} = \mathbb{E}_{z, c, t, \epsilon, c_{RM}}[\|\epsilon - \epsilon_{\theta}(z_t, c, t, c_{RM})\|_{2}^{2}],
\end{equation}
where $z_t$ denotes the latent features at timestep $t$, $\epsilon$ is the ground-truth Gaussian noise, and $\epsilon_\theta$ represents the denoising network. The term $c$ refers to the text condition, while $c_{RM}$ serves as the structural control condition derived from the RM output. By treating the adversarial trigger~\cite{song2026segtrans} as \textit{out-of-distribution} noise, this process selectively eliminates it while faithfully re-synthesizing the sharp edges required for the detector's regression head. Formally, we encapsulate this two-stage restoration process into a unified reconstruction operator $\mathcal{R}(\cdot)$. For each corrupted variant $I_{corr} \in \mathcal{I}_{corr}$, the model yields a purified counterpart $I_{purified} = \mathcal{R}(I_{corr})$, effectively resolving the trade-off between attack disruption and multi-object preservation.

\subsection{Selection: Distilling High-Confidence Detections}\label{sec:selection_methodology}
Detectors like Faster R-CNN operate on the principle of dense prediction. In our CRS paradigm, the diverse corrupted variants compel the detector to produce a massive pool of redundant predictions. Leveraging the \textit{robustness asymmetry} identified in Section~\ref{4-1}, we exploit the fact that true positive detections exhibit spatial consistency (stemming from the \textit{consistent vulnerability} of benign features), whereas backdoor-induced anomalies (\textit{e.g.,} false positives in OGA) exhibit significant instability (reflecting the \textit{variable vulnerability} of triggers) across corrupted variants. To distill robust detections based on this principle, we propose a multi-stage process that acts as \textit{Cross-Corruption Non-Maximum Suppression}.

\noindent \textbf{Candidate Box Aggregation.}
First, we aggregate all detection results from the purified variants into a comprehensive set $\mathcal{B}_{cand}$. Each detection is a tuple $\hat{r} = (\hat{b}, \hat{l}, \hat{s})$, comprising a bounding box $\hat{b}$, a class label $\hat{l}$, and a confidence score $\hat{s}$. To facilitate spatial clustering, we normalize the center coordinates $(\hat{c}_{i,x}, \hat{c}_{i,y})$ of box $\hat{b}_i$ by the image dimensions $(W, H)$:
\begin{equation}
    \hat{c}_{i,x} = \frac{(\hat{b}_i^{x_1} + \hat{b}_i^{x_2})}{2W}, \quad \hat{c}_{i,y} = \frac{(\hat{b}_i^{y_1} + \hat{b}_i^{y_2})}{2H}.
\end{equation}
To ensure semantic consistency, we partition $\mathcal{B}_{cand}$ into class-specific subsets $\mathcal{B}_{\hat{l}} = \{(\hat{b}, \hat{l}', \hat{s}) \in \mathcal{B}_{cand} \mid \hat{l}' = \hat{l}\}$.

\noindent \textbf{Composite Distance Metric.}
To quantify the similarity between candidate detections robustly against regression jitter, we construct a Composite Distance Metric. For any two detections $\hat{r}_i, \hat{r}_j \in \mathcal{B}_{\hat{l}}$, their composite distance $D(\hat{r}_i, \hat{r}_j)$ is evaluated over their bounding boxes $\hat{b}_i, \hat{b}_j$ as:
\begin{IEEEeqnarray}{rCl}
    D(\hat{r}_i, \hat{r}_j) & = & d_1 + d_2.
\end{IEEEeqnarray}
Here, $d_1$ measures the overlap between the two boxes and is derived from the IoU as follows:
\begin{IEEEeqnarray}{rCl}
    d_1 & = & 1 - \text{IoU}(\hat{b}_i, \hat{b}_j), \quad d_1 \in [0, 1].
\end{IEEEeqnarray}
The component $d_2$ quantifies the spatial proximity of the box centers using a normalized Euclidean distance:
\begin{IEEEeqnarray}{rCl}
    d_2 & = & \sqrt{\frac{(\hat{c}_{i,x} - \hat{c}_{j,x})^2 + (\hat{c}_{i,y} - \hat{c}_{j,y})^2}{2}}, \quad d_2 \in [0, 1].
\end{IEEEeqnarray}
A smaller value of $D(\hat{r}_i, \hat{r}_j)$ corresponds to a higher degree of spatial similarity between the two candidate detections.

\noindent \textbf{Bounding Box Clustering.}
For each class-specific subset $\mathcal{B}_{\hat{l}}$, we apply the DBSCAN algorithm~\cite{Ester1996DBSCAN} using metric $D$, governed by a distance threshold $\epsilon$ and a minimum cluster size $\text{MinPts}$. The objective is to partition $\mathcal{B}_{\hat{l}}$ into distinct spatial clusters $\mathcal{K}_{\hat{l}} = \{K_1, \ldots, K_{k_{\hat{l}}}\}$ and a set of noise detections $\mathcal{N}_{\hat{l}}$:
\begin{IEEEeqnarray}{rCl}
    \mathcal{B}_{\hat{l}} & = & \left( \bigcup_{m=1}^{k_{\hat{l}}} K_m \right) \cup \mathcal{N}_{\hat{l}}.
\end{IEEEeqnarray}
This clustering process essentially identifies the consensus across diverse corrupted variants. Here, $\text{MinPts}$ serves as the \textit{consensus threshold}, determining the minimum number of consistent detections required to validate an object. Since triggers are sensitive to input perturbations (\textit{variable vulnerability}), their induced flickering outliers fail to form dense clusters, signifying a lack of consensus, and are thus discarded as noise $\mathcal{N}_{\hat{l}}$. Conversely, benign objects exhibit spatial consistency, forming stable clusters that represent a strong consensus.

\noindent \textbf{Confidence-Based Voting.}
Finally, for each valid cluster $K_m \in \mathcal{K}_{\hat{l}}$ (representing an agreed-upon physical target), we perform confidence-based voting to elect the most reliable representative. We designate the detection tuple with the highest confidence score $\hat{s}$ within the cluster as the final detection $\hat{r}_{final}^{(\hat{l}, m)}$:
\begin{IEEEeqnarray}{rCl}
    \hat{r}_{final}^{(\hat{l}, m)} & = & \argmax_{(\hat{b}, \hat{l}', \hat{s}) \in K_m} \ \hat{s}.
\end{IEEEeqnarray}
Aggregating the elected representatives across all unique predicted classes $\mathcal{L}_{unique}$ systematically distills the noisy candidate pool into the final high-confidence detection set $\mathcal{B}_{final} = \bigcup_{\hat{l} \in \mathcal{L}_{unique}} \{ \hat{r}_{final}^{(\hat{l}, m)} \}_{m=1}^{k_{\hat{l}}}$, effectively materializing the ensemble corruption consensus into precise predictions. The complete workflow of ODPure is summarized in Algorithm~\ref{alg:odpure}.

\begin{algorithm}[hbt!]
\caption{The Complete ODPure Scheme}
\label{alg:odpure}
\begin{algorithmic}[1]
\Require
    $I$: An input image (potentially malicious).
    $M(\cdot)$: A black-box object detector.
    $\mathcal{F}$: A portfolio of corruption functions $\{f_{i,j}\}$.
    $\mathcal{R}(\cdot)$: The high-quality image reconstruction function.
    $\epsilon, \text{MinPts}$: Parameters for DBSCAN.
\Ensure
    $\mathcal{B}_{final}$: A set of purified, high-confidence detections.
\State Initialize $\mathcal{B}_{cand} \gets \emptyset$, $\mathcal{B}_{final} \gets \emptyset$

\For{each corruption function $f \in \mathcal{F}$}
    \State $I_{corr} \gets f(I)$
    \State $I_{purified} \gets \mathcal{R}(I_{corr})$
    \State $\mathcal{B} \gets M(I_{purified})$ \Comment{Detections $\{(\hat{b}, \hat{l}, \hat{s})\}$}
    \State $\mathcal{B}_{cand} \gets \mathcal{B}_{cand} \cup \mathcal{B}$
\EndFor

\State $\mathcal{L}_{unique} \gets \{\hat{l} \mid (\hat{b}, \hat{l}, \hat{s}) \in \mathcal{B}_{cand}\}$ 

\For{each class label $\hat{l} \in \mathcal{L}_{unique}$}
    \State $\mathcal{B}_{\hat{l}} \gets \{(\hat{b}, \hat{l}', \hat{s}) \in \mathcal{B}_{cand} \mid \hat{l}' = \hat{l}\}$
    \If{$|\mathcal{B}_{\hat{l}}| \ge \text{MinPts}$}
        \State $D_M \gets \text{ComputeDistanceMatrix}(\mathcal{B}_{\hat{l}})$
        \State $\mathcal{K}_{\hat{l}}, \mathcal{N}_{\hat{l}} \gets \text{DBSCAN}(D_M, \epsilon, \text{MinPts})$
        \For{each cluster $K_m \in \mathcal{K}_{\hat{l}}$}
            \State $\hat{r}_{final} \gets \argmax_{(\hat{b}, \hat{l}', \hat{s}) \in K_m} \ \hat{s}$
            \State $\mathcal{B}_{final} \gets \mathcal{B}_{final} \cup \{\hat{r}_{final}\}$
        \EndFor
    \EndIf
\EndFor

\State \Return $\mathcal{B}_{final}$
\end{algorithmic}
\end{algorithm}

\section{Experiments}\label{sec22}
This section describes the datasets, models, and configurations used in our experiments on attacks and defenses, and presents a series of evaluation results.
\subsection{Experimental Setup}
\noindent \textbf{Dataset.} We evaluate our method on two widely used real-world object detection benchmark datasets: COCO~\cite{Lin2014COCO} (2017) and VOC~\cite{Everingham2010VOC} (2007+2012 union).

\begin{figure}[!t]
    \centering
    
    \begin{subfigure}[b]{0.25\columnwidth}
        \centering
        \includegraphics[width=\linewidth]{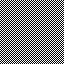}
        \caption{Chessboard.}
        \label{fig:trigger_chessboard}
    \end{subfigure}%
    \hspace{0.05\columnwidth}%
    \begin{subfigure}[b]{0.25\columnwidth}
        \centering
        \fbox{\includegraphics[width=0.94\linewidth, height=0.94\linewidth]{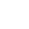}}
        \caption{Solid White.}
        \label{fig:trigger_solid}
    \end{subfigure}%
    \hspace{0.05\columnwidth}%
    \begin{subfigure}[b]{0.25\columnwidth}
        \centering
        \includegraphics[width=\linewidth]{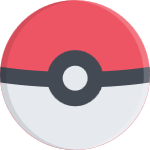}
        \caption{Pok\'{e} Ball.}
        \label{fig:trigger_pokeball}
    \end{subfigure}

    \caption{The trigger patterns.}
    \label{fig:trigger_examples}
    \vspace{-2mm}
\end{figure}

\noindent \textbf{Configurations for Backdoor Attacks.}
We follow~\cite{Chan2022Baddet} for injecting a backdoor during training. We select a target label \textit{person} and poison 10--30\% of the training set, which maintains both a high ASR and clean performance. Unless otherwise specified, we utilize a chessboard trigger (Fig.~\ref{fig:trigger_chessboard}) sized $29\times29$ or $15\times15$ pixels, balancing mAP and ASR in all experiments. Detailed configurations for different attack methods and datasets are provided in Section~\ref{secA} of the Supplementary Material.

\noindent \textbf{Configuration for ODPure.} Our implementation follows the CRS paradigm (Section~\ref{5}). For the \textit{Corruption} stage (Section~\ref{5-1}), we use the $n=15$ corruption types benchmarked by~\cite{Liu2023Detecting} (see Section~\ref{secA} of the Supplementary Material for the full list), applied across $m=3$ severity levels (1, 2, and 3). For the \textit{Reconstruction} stage, we adopt the two-stage modified architecture from Section~\ref{5-2}, with model configurations identical to those in~\cite{Lin2024DiffBIR}. For the \textit{Selection} stage, we utilize the algorithm in Section~\ref{sec:selection_methodology} with the distance threshold $\epsilon$ set to 0.5 and the core parameter $\text{MinPts}$ set to 10.

\noindent \textbf{Models.} We evaluate our method on two representative object detectors: the single-stage YOLOv5~\cite{Redmon2016YOLO, Jocher2021YOLOv5} and the two-stage Faster R-CNN~\cite{Ren2017Faster}.

\noindent \textbf{Metrics.} We report mAP and ASR (see Section~\ref{sec3-2}) before and after our defense to measure not only the reduction in ASR but also to understand the impact of ODPure on performance.

\noindent \textbf{Competitors.} As the pioneering input-stage purification defense for object detection, ODPure lacks direct counterparts within the same stage. To ensure a comprehensive evaluation despite this absence, we benchmark against the most relevant methods across different paradigms (summarized in Section~\ref{6-5}). Specifically, we first qualitatively compare against detector-specific defenses at alternative stages (DetectorCleanse~\cite{Chan2022Baddet} and ODSCAN~\cite{Cheng2024ODScan}). Furthermore, to establish a direct quantitative baseline, we adapt the state-of-the-art classification-based purification method, ZIP~\cite{Shi2023ZIP}, to object detection.

\subsection{Results of ODPure for OMA, ODA, OGA}\label{6-2}
In this subsection, we evaluate the efficacy of our defense against three representative backdoor attacks in object detection~\cite{Chan2022Baddet, Cheng2024ODScan, Shen2023Django, Luo2023Untargeted, TrojAI-leaderboard}: OMA, ODA, and OGA, with results summarized in Table~\ref{tab:combined_attacks}. The effectiveness of our defense is gauged by two key metrics: its ability to significantly reduce the ASR, and its capacity to preserve a high mAP on clean samples.

\begin{table}[htbp]
\centering
\caption{Robustness against three representative backdoor attacks (OMA, ODA, OGA) on VOC and COCO datasets.}
\label{tab:combined_attacks}
\scriptsize
\setlength{\tabcolsep}{0pt} 
\renewcommand{\arraystretch}{1.1}

\begin{tabular*}{\columnwidth}{@{\extracolsep{\fill}} l l 
    c @{\hspace{1.0em}} 
    c @{\hspace{0.7em}} c 
    @{\hspace{0.2em}} 
    c @{\hspace{0.7em}} c 
    @{}}
\toprule
\multirow{2}{*}{\textbf{Attack}} & 
\multirow{2}{*}{\textbf{Dataset (Model)}} & 
\textbf{Clean} & 
\multicolumn{2}{c}{\textbf{Before Defense}} & 
\multicolumn{2}{c}{\textbf{After Defense}} \\
\cmidrule(r){3-3} \cmidrule{4-5} \cmidrule{6-7}
& & mAP & mAP & ASR & mAP & ASR \\
\midrule
\multirow{4}{*}{OMA} 
& VOC (YOLO)   & 76.4\% & 8.2\%  & 87.7\% & \textbf{80.5\%} & \textbf{2.0\%} \\
& VOC (F-RCNN) & 79.3\% & 44.9\% & 94.6\% & 78.1\%          & \textbf{17.4\%} \\
& COCO (YOLO)  & 52.8\% & 0.4\%  & 94.6\% & 52.0\%          & \textbf{1.5\%} \\
& COCO (F-RCNN)& 49.7\% & 6.3\%  & 91.9\% & 47.0\%          & \textbf{16.3\%} \\
\midrule
\multirow{4}{*}{ODA} 
& VOC (YOLO)   & 72.0\% & 71.6\% & 96.5\% & \textbf{76.7\%} & \textbf{20.8\%} \\
& VOC (F-RCNN) & 77.6\% & 76.4\% & 69.3\% & 75.4\%          & \textbf{18.9\%} \\
& COCO (YOLO)  & 54.0\% & 52.4\% & 99.9\% & \textbf{54.1\%} & \textbf{25.4\%} \\
& COCO (F-RCNN)& 50.9\% & 50.3\% & 81.7\% & \textbf{51.4\%} & \textbf{28.0\%} \\
\midrule
\multirow{4}{*}{OGA} 
& VOC (YOLO)   & 80.4\% & 78.0\% & 65.1\% & \textbf{82.2\%} & \textbf{0.0\%} \\
& VOC (F-RCNN) & 83.2\% & 81.2\% & 98.4\% & 80.9\%          & \textbf{0.0\%} \\
& COCO (YOLO)  & 53.0\% & 52.8\% & 99.8\% & \textbf{54.4\%} & \textbf{0.0\%} \\
& COCO (F-RCNN)& 48.9\% & 49.1\% & 95.4\% & \textbf{49.5\%} & \textbf{0.0\%} \\
\bottomrule
\end{tabular*}
\end{table}

\noindent \textbf{Object Misclassification Attack.}
In OMA, the backdoor is designed to force the detector to misclassify all non-target objects as a predefined target class (\textit{e.g.,} \textit{person}) whenever the trigger is present. Consequently, this manipulation floods the output with false positives for the target category while suppressing legitimate objects into false negatives, causing the mAP to collapse. This effect is illustrated in Table~\ref{tab:combined_attacks}. For the YOLO model, the attack is devastating: the mAP on VOC plummets from a clean baseline of 76.4\% to a mere 8.2\%, and this trend is even more pronounced on the more challenging COCO dataset, where the mAP drops to nearly zero (\textbf{0.4\%}).

In the face of this severe performance degradation, ODPure demonstrates remarkable efficacy. For the YOLO model on VOC, it not only neutralizes the ASR to just \textbf{2.0\%} but restores the mAP to \textbf{80.5\%}, fully recovering and even slightly exceeding its original performance. This robust recovery is replicated on COCO, where ODPure restores the mAP from 0.4\% back to 52.0\% while neutralizing the ASR to \textbf{1.5\%}. We attribute this slight performance gain to the image enhancement effect of our \textit{Reconstruction} stage. By leveraging generative priors, this stage not only eradicates triggers but also repairs native image degradations (\textit{e.g.,} compression artifacts) present in the original datasets, thereby sharpening semantic features for the detector.

\noindent \textbf{Object Disappearance Attack.}
ODA renders a specific target class of objects invisible to the detector, erasing them from the output~\cite{Chan2022Baddet, Cheng2024ODScan}. The results in Table~\ref{tab:combined_attacks} confirm the efficacy of this attack: ODA achieves a near-total ASR (\textit{e.g.,} 96.5\% for YOLO on VOC) while strategically preserving the model's overall utility. This is evidenced by the minimal degradation in mAP on the poisoned model compared to the clean baseline (a mere drop from 72.0\% to 71.6\%). This controlled impact makes detecting the attack challenging, as it avoids causing a catastrophic failure that would be easily flagged.

A successful defense must not only reduce ASR but critically restore mAP by recovering disappeared objects. Despite the attack's design, ODPure proves to be a robust countermeasure. On the VOC dataset (Table~\ref{tab:combined_attacks}), it reduces the ASR to \textbf{20.8\%} while fully restoring and enhancing the mAP to \textbf{76.7\%}.

\noindent \textbf{Object Generation Attack.}
OGA forces the detector to generate a false-positive bounding box of a predefined size ($W \times L$) and target class, centered on the trigger's location. Table~\ref{tab:combined_attacks} confirms this efficacy. OGA achieves a high ASR while strategically preserving the detector's performance on legitimate objects, causing only a minimal drop in mAP (\textit{e.g.,} from 80.4\% to 78.0\% for YOLO on VOC).

Despite this, ODPure proves to be a definitive countermeasure. Across all tested models and datasets, our method \textbf{completely nullifies the attack, reducing the ASR to a perfect 0.0\%}. More impressively, the purification process does more than simply maintain the model's performance; it consistently enhances the mAP beyond the original clean baseline. For instance, on the COCO dataset, ODPure boosts the YOLO model's mAP from 52.8\% to \textbf{54.4\%}, surpassing its clean performance of 53.0\%. This recurrence of performance gain further corroborates the beneficial impact of our reconstruction stage in enhancing image quality as discussed earlier.

\subsection{Comparing Detector Defenses}\label{6-5}

Table~\ref{tab:qualitative_comp} qualitatively compares ODPure's architectural distinctiveness. Unlike post-inference DetectorCleanse~\cite{Chan2022Baddet} or pre-deployment ODSCAN~\cite{Cheng2024ODScan}, which rely on a detect-and-discard paradigm, ODPure purifies at the input stage. This fundamental difference allows ODPure to circumvent the white-box requirements of model scanning and, more critically, avoid the discard-based limitation of output-scanning approaches. By enabling the continued use of sanitized inputs, ODPure ensures the reusability of both the data and the detector, establishing itself as a practical and non-destructive line of defense.

The quantitative results of this comparison are presented in Table~\ref{tab:zip_results}. In object detection, merely reducing the ASR is insufficient; preserving the model's mAP is paramount, as it reflects the critical ability to accurately localize and classify all objects in a scene. This trade-off is highlighted starkly by our results. The ZIP method, in its attempt to eliminate the trigger, causes a catastrophic drop in detection accuracy, with the mAP plummeting from 52.8\% to \textbf{30.6\%}. In contrast, ODPure achieves a superior ASR reduction (1.5\% vs. 2.8\% for ZIP) while effectively preserving the mAP at 52.0\%. This demonstrates that our CRS paradigm, designed specifically for the complexities of object detection, successfully purifies the input without sacrificing the detector's core performance.

\begin{table}[h!]
\centering
\caption{Comparing ODPure with other detector defenses.}
\label{tab:qualitative_comp}
\small
\renewcommand{\arraystretch}{1.2}
\newcolumntype{L}{>{\raggedright\arraybackslash}X}
\renewcommand{\tabularxcolumn}[1]{m{#1}}
\begin{tabularx}{\columnwidth}{@{} 
    >{\hsize=1.0\hsize\raggedright\arraybackslash}X
    >{\hsize=0.8\hsize\centering\arraybackslash}X
    >{\hsize=0.9\hsize\centering\arraybackslash}X
    >{\hsize=1.3\hsize\centering\arraybackslash}X
@{}}
\toprule
\mbox{\textbf{Work}} & 
\mbox{\textbf{Type}} & 
\mbox{\textbf{Defense Scope}} & 
\mbox{\textbf{Reusability}} \\
\midrule
ODSCAN~\cite{Cheng2024ODScan} & 
Model scanning & 
All & 
\ding{55} \par (Discard detectors) \\
DetectorCleanse~\cite{Chan2022Baddet} & 
Output scanning & 
All & 
\ding{55} \par (Discard inputs \& disrupts service) \\
\textbf{ODPure (Ours)} & 
Input purification & 
All & 
\checkmark \par (Detector always operational) \\
\bottomrule
\end{tabularx}
\end{table}

\begin{table}[htbp]
\centering
\caption{Performance comparison between ODPure and the classification-based purification method ZIP~\cite{Shi2023ZIP} on the COCO dataset, using the OMA-attacked YOLO model.}
\label{tab:zip_results}
\scriptsize
\setlength{\tabcolsep}{0pt}

\begin{tabular*}{\columnwidth}{@{\extracolsep{\fill}} l
    c @{\hspace{1.0em}} c
    @{\hspace{0.2em}}
    c @{\hspace{0.7em}} c
    @{}}
\toprule
\multirow{2}{*}{\textbf{Method}}
& \multicolumn{2}{c}{\textbf{\shortstack[c]{\centering Before Defense}}}
& \multicolumn{2}{c}{\textbf{\parbox{1.8cm}{\centering After Defense}}} \\
\cmidrule{2-3} \cmidrule{4-5}
& mAP & ASR & mAP & ASR \\
\midrule
ZIP    & 52.8\% & 94.6\% & 30.6\% & 2.8\% \\
ODPure & 52.8\% & 94.6\% & \textbf{52.0\%} & \textbf{1.5\%} \\
\bottomrule
\end{tabular*}
\end{table}

\subsection{Adaptive Attacks}\label{6-3}
We also evaluate ODPure against an adaptive attacker who, with knowledge of our defense mechanism, attempts to craft evasion triggers. We consider two primary  adaptive strategies:

\begin{itemize}
    \item The first strategy employs a simple, solid color trigger such as a solid white patch (see Fig.~\ref{fig:trigger_solid}). The hypothesis is that such low-frequency, minimalist patterns may be more resilient to the diverse corruptions applied in our first stage, challenging the efficacy of the \textit{Corruption} phase.
    \item The second strategy utilizes triggers that mimic real-world objects (\textit{i.e.,} physical-world triggers), such as a Pok\'{e} Ball sticker (see Fig.~\ref{fig:trigger_pokeball}). This approach exploits the generative prior of our \textit{Reconstruction} module, potentially tricking it into restoring the trigger as a benign object.
\end{itemize}

In summary, while real-world triggers (Fig.~\ref{fig:trigger_pokeball}) can marginally increase the ASR, ODPure remains highly effective by significantly reducing their ASR, as detailed in Table~\ref{tab:adaptive_physical}. This efficacy is largely attributed to the intensity and diversity of our \textit{Corruption} phase, which sufficiently disrupts the trigger's features to prevent its faithful reconstruction. In contrast, solid color triggers pose a greater challenge, leading to a more noticeable performance degradation. Nevertheless, even against this more potent attack, our method reduces the ASR by over 60\% while maintaining high mAP. The complete results for the solid color trigger and detailed analyses of these strategies are provided in Section~\ref{secB} of the Supplementary Material.

\begin{table}[htbp]
\centering
\caption{Robustness against adaptive attacks with the Physical-world Trigger (Pok\'{e} Ball).}
\label{tab:adaptive_physical}
\scriptsize
\setlength{\tabcolsep}{0pt}
\renewcommand{\arraystretch}{1.1}

\begin{tabular*}{\columnwidth}{@{\extracolsep{\fill}} l l 
    c @{\hspace{1.0em}} 
    c @{\hspace{0.7em}} c 
    @{\hspace{0.2em}} 
    c @{\hspace{0.7em}} c 
    @{}}
\toprule
\multirow{2}{*}{\textbf{Attack}} & 
\multirow{2}{*}{\textbf{Dataset (Model)}} & 
\textbf{Clean} & 
\multicolumn{2}{c}{\textbf{Before Defense}} & 
\multicolumn{2}{c}{\textbf{After Defense}} \\
\cmidrule(r){3-3} \cmidrule{4-5} \cmidrule{6-7}
& & mAP & mAP & ASR & mAP & ASR \\
\midrule
\multirow{4}{*}{OMA} 
& VOC (YOLO)   & 77.3\% & 16.1\% & 95.5\% & \textbf{78.3\%} & \textbf{19.8\%} \\
& VOC (F-RCNN) & 80.5\% & 43.7\% & 95.1\% & 78.1\%          & \textbf{46.0\%} \\
& COCO (YOLO)  & 54.5\% & 1.7\%  & 95.4\% & 51.2\%          & \textbf{31.1\%} \\
& COCO (F-RCNN)& 49.4\% & 8.1\%  & 90.9\% & \textbf{50.5\%} & \textbf{34.9\%} \\
\midrule
\multirow{4}{*}{ODA} 
& VOC (YOLO)   & 75.3\% & 72.3\% & 98.5\% & \textbf{78.3\%} & \textbf{35.1\%} \\
& VOC (F-RCNN) & 78.3\% & 77.3\% & 57.7\% & 76.2\%          & \textbf{17.8\%} \\
& COCO (YOLO)  & 55.3\% & 53.2\% & 99.5\% & 52.7\%          & \textbf{26.2\%} \\
& COCO (F-RCNN)& 51.8\% & 52.5\% & 76.1\% & \textbf{52.4\%} & \textbf{21.3\%} \\
\midrule
\multirow{4}{*}{OGA} 
& VOC (YOLO)   & 79.8\% & 77.0\% & 96.8\% & \textbf{81.4\%} & \textbf{15.4\%} \\
& VOC (F-RCNN) & 84.1\% & 80.8\% & 99.6\% & 80.3\%          & \textbf{6.7\%} \\
& COCO (YOLO)  & 54.0\% & 52.3\% & 99.8\% & 52.8\%          & \textbf{29.2\%} \\
& COCO (F-RCNN)& 51.4\% & 50.3\% & 97.6\% & \textbf{51.8\%} & \textbf{12.0\%} \\
\bottomrule
\end{tabular*}
\end{table}

\subsection{Ablation Study}
\label{sec:ablation}

We analyze the contributions of each component in our CRS paradigm. Unless otherwise specified, all experiments in this subsection are performed on the ODA-attacked YOLO detector using the chessboard trigger of $29\times29$ pixels. Detailed baseline configurations and experimental settings are provided in Section~\ref{secA} of the Supplementary Material.

\begin{figure}[!t]
    \centering
    \subfloat[\footnotesize ASR after corruption.]{\includegraphics[width=0.46\columnwidth]{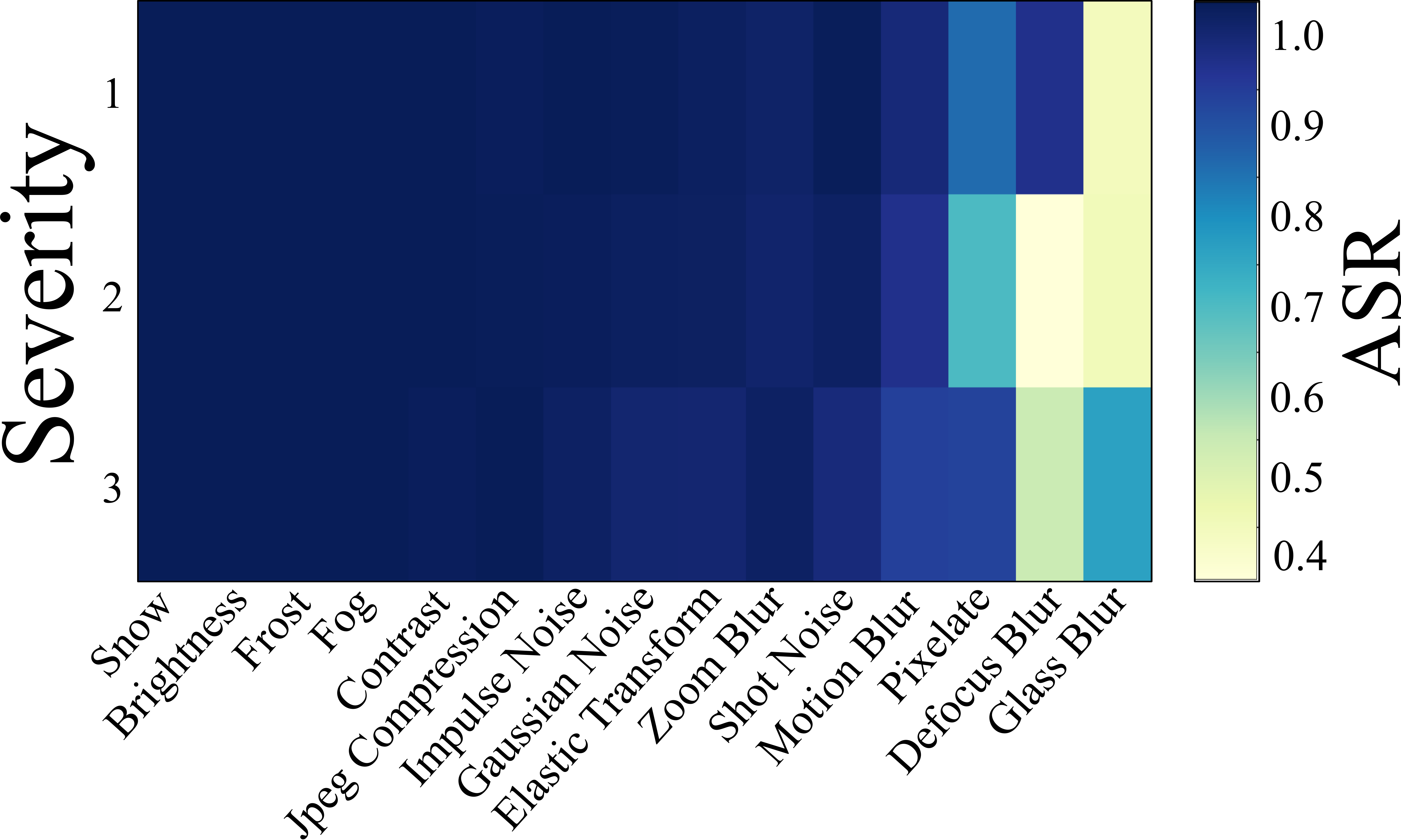}\label{fig:heatmap_asr}} \qquad
    \subfloat[\footnotesize mAP after corruption.]{\includegraphics[width=0.46\columnwidth]{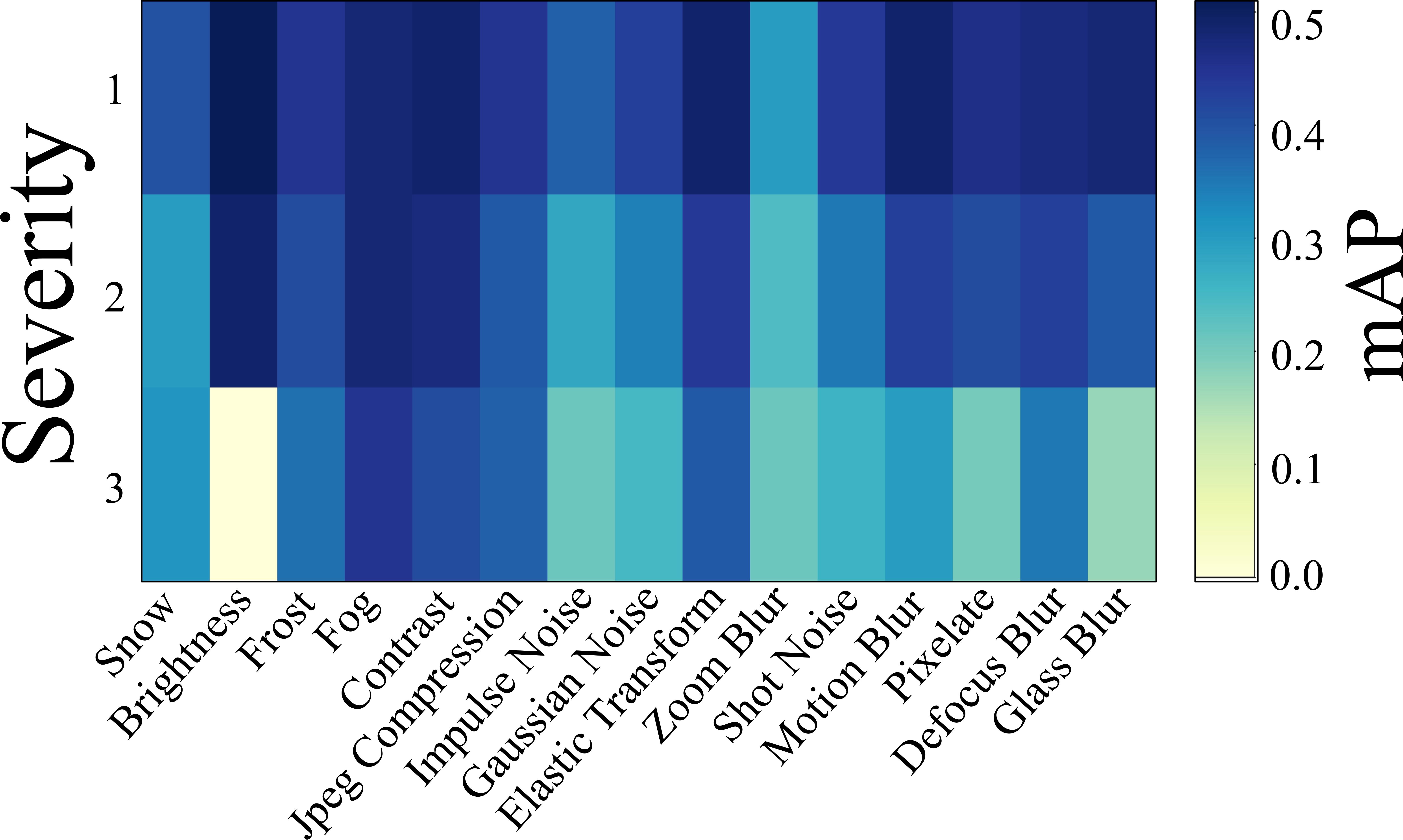}\label{fig:heatmap_map}}
    
    \captionsetup{font=small}
    \caption{Performance analysis of a backdoored YOLO model on the poisoned COCO dataset after applying only the \textit{Corruption} step. The heatmaps visualize (a) ASR and (b) mAP across 15 corruption types and 3 severity levels.}
    \label{fig:heatmap_comparison}
    \vspace{-2mm}
\end{figure}

\noindent \textbf{Trade-offs of a Corruption-Only Defense.} We first analyze the impact of the \textit{Corruption} module in isolation across 15 corruption types and three severities, as visualized in Fig.~\ref{fig:heatmap_comparison}.

Our findings reveal a critical trade-off. While specific corruptions like Defocus Blur (severity 2) can effectively disrupt the trigger (reducing ASR to 34.5\%), they severely degrade the mAP. Conversely, other corruptions (\textit{e.g.,} snow) preserve mAP but negligibly affect ASR. This indicates that relying on any single corruption type is not a robust strategy.

This demonstrates that the \textit{Corruption} phase, while necessary to disrupt the trigger, is insufficient alone as it severely compromises model utility. Consequently, it highlights the need for the \textit{Reconstruction} phase to restore the features of benign objects and the \textit{Selection} phase to distill a robust prediction.

\noindent \textbf{Necessity of Corruption Diversity.}
To isolate the effect of corruption diversity, we evaluate a simplified variant using only the single most effective corruption (defocus blur, severity 2).  We assess this setup using YOLO on both VOC and COCO.

Table~\ref{tab:combined_ablation} clearly highlights the necessity of employing a diverse corruption set. While the simplified ODPure variant utilizing even the strongest single corruption manages to reduce the ASR to some extent (\textit{i.e.,} from 96.5\% to 37.4\% on VOC), it fails to completely neutralize the attack and offers suboptimal mAP improvements compared to the full ODPure.

\begin{table}[htbp]
\centering
\caption{Ablation study on the three stages of the CRS paradigm using the ODA-attacked YOLO model.}
\label{tab:combined_ablation}
\scriptsize
\setlength{\tabcolsep}{3pt}
\renewcommand{\arraystretch}{1.1}

\begin{tabular*}{\columnwidth}{@{\extracolsep{\fill}}l cc cc}
\toprule
\multirow{2}{*}{\textbf{Configuration}} & \multicolumn{2}{c}{\textbf{VOC}} & \multicolumn{2}{c}{\textbf{COCO}} \\
\cmidrule{2-3} \cmidrule{4-5}
& mAP & ASR & mAP & ASR \\
\midrule
Undefended Baseline & 71.6\% & 96.5\% & 52.4\% & 99.9\% \\
\midrule
\textbf{Stage 1: Corruption} & & & & \\
\quad ODPure (Single Corruption) & 72.8\% & 37.4\% & 51.5\% & 30.0\% \\
\midrule
\textbf{Stage 2: Reconstruction} & & & & \\
\quad ODPure (w/o Reconstruction) & 51.5\% & 88.2\% & 51.4\% & 94.5\% \\
\midrule
\textbf{Stage 3: Selection} & & & & \\
\quad ODPure (Random Selection) & 69.1\% & 47.7\% & 45.9\% & 38.4\% \\
\midrule
\textbf{ODPure (Full Method)} & \textbf{76.7\%} & \textbf{20.8\%} & \textbf{54.1\%} & \textbf{25.4\%} \\
\bottomrule
\end{tabular*}
\end{table}

In stark contrast, our full ODPure (using all 15 types) performs significantly better. On VOC, it reduces the ASR more effectively (to \textbf{20.8\%}) and boosts mAP from 71.6\% to \textbf{76.7\%}. A similar trend is observed on COCO. This demonstrates that corruption diversity is crucial for the CRS paradigm; it challenges trigger survival, enabling the subsequent \textit{Reconstruction} and \textit{Selection} phases to better distinguish and eliminate backdoor-induced artifacts while restoring high-fidelity legitimate object features for detection.

\noindent \textbf{Necessity of Reconstruction.}
To validate the necessity of the \textit{Reconstruction} module, we evaluate a variant where corrupted samples are fed directly to the \textit{Selection} phase.

Table~\ref{tab:combined_ablation} demonstrates that the \textit{Reconstruction} module is indispensable. Omitting it (the ODPure (w/o Reconstruction) variant) renders the defense ineffective. Although \textit{Corruption} and \textit{Selection} alone slightly reduce ASR, it remains unacceptably high (88.2\% on VOC and 94.5\% on COCO). Critically, the model's overall utility is severely compromised, with the mAP plummeting to 51.5\% on both datasets, indicating that without reconstruction, aggressive corruption irreparably damages legitimate object features.

Conversely, our full ODPure method yields significant improvements. It not only reduces the ASR to \textbf{20.8\%} on VOC but also successfully restores the mAP to its original level (from a degraded 51.5\% back to a robust \textbf{76.7\%}). This confirms that our \textit{Reconstruction} phase plays a crucial dual role: it restores benign object features to preserve utility and further suppresses disrupted trigger patterns.

\noindent \textbf{Necessity of Consensus-Based Selection.}
Finally, to validate our consensus-based selection mechanism, we evaluate a variant where the selection module is replaced by a random sampling strategy (picking one of the 45 purified results randomly).

Table~\ref{tab:combined_ablation} underscores the critical role of our strategy. While random sampling lowers ASR compared to the undefended model, it comes at a significant cost to utility, causing mAP to drop noticeably (\textit{i.e.,} from 71.6\% to 69.1\% on VOC and from 52.4\% to 45.9\% on COCO). This indicates that random selection is unreliable, as it often picks a sample where legitimate objects have been overly corrupted.

Comparatively, our full ODPure method, equipped with the clustering and voting mechanism, achieves superior ASR reduction and consistently preserves or even enhances mAP. This improvement is likely attributable to the ensemble-like effect of our strategy, which capitalizes on the implicit data augmentation from the \textit{Reconstruction} process. This demonstrates that our selection process effectively distills a robust, high-confidence prediction from the diverse set of purified results, filtering out noise while consolidating the detections of true objects.

\noindent \textbf{Inference Efficiency.} We evaluated the computational cost on a single NVIDIA RTX 3090 GPU. ODPure requires approximately $15$ seconds per image, with the latency dominated by the iterative sampling of the diffusion-based \textit{Reconstruction} stage, while corruption and selection incur negligible overhead ($<0.05$s). This overhead is immediately practical for high-stakes scenarios such as cloud-assisted perception verification, dataset sanitization, and offline safety audits where purification fidelity is paramount. Furthermore, this latency can be drastically curtailed by integrating step-distilled generative priors (\textit{e.g.,} Latent Consistency Models~\cite{Luo2023LCM}), providing a viable trajectory toward real-time onboard edge deployment without altering the CRS paradigm.

\section{Conclusion}
In this paper, we addressed the critical challenge of backdoor defense for object detectors by proposing ODPure, the first input-stage, black-box purification method. Leveraging a novel \textit{Corruption-Reconstruction-Selection} paradigm, it effectively neutralizes diverse backdoor attacks while preserving high model utility, filling a critical gap by enabling data reusability where existing discard-based detection defenses fall short. Extensive experiments validated ODPure's state-of-the-art performance and robustness, even against adaptive attacks. Regarding limitations, the current multi-stage pipeline entails computational latency largely driven by iterative diffusion sampling. Future work will focus on integrating algorithmic acceleration, such as latent consistency distillation and efficient numerical solvers, to reduce inference steps by an order of magnitude without compromising purification fidelity.

\bibliographystyle{IEEEtran}
\bibliography{ref}


\clearpage 

\setcounter{section}{0}
\setcounter{figure}{0}
\setcounter{table}{0}
\setcounter{equation}{0}

\renewcommand{\thesection}{S-\Roman{section}}
\renewcommand{\thefigure}{S\arabic{figure}}
\renewcommand{\thetable}{S\arabic{table}}
\renewcommand{\theequation}{S-\arabic{equation}}

\twocolumn[
  \begin{@twocolumnfalse}
    \centering
    \vspace{0.5em}
    {\LARGE \textbf{Supplementary Materials for \\ ``ODPure: Backdoor Purification for Object Detection \\ via Ensemble Corruption Consensus''}\par}
    \vspace{1.5em}
  \end{@twocolumnfalse}
]

\section{Detailed Configurations for Experiments}\label{secA}

In our experiments, the backdoor attack parameters and training settings for YOLO and Faster R-CNN primarily follow the methodology in~\cite{Chan2022Baddet}, with a uniform trigger transparency of 0.5. We configure three distinct attack types: Object Misclassification Attack (OMA), Object Disappearance Attack (ODA), and Object Generation Attack (OGA). To ensure a fair comparison, we maintain consistent poisoning rates across both standard and adaptive attack scenarios: 30\% for OMA, 20\% for ODA, and 10\% for OGA.

Regarding trigger dimensions, configurations differ slightly between scenarios. For standard attacks (Chessboard), we use a patch of $29\times29$ pixels for OMA and ODA, and a patch of $15\times15$ pixels for OGA. For adaptive attacks (Solid White and Pok\'{e} Ball), a uniform trigger of $15\times15$ pixels is applied across all three attack types. Additionally, for OGA, the generated false bounding box is fixed at $30\times60$ pixels.

\noindent \textbf{Corruption Set Details.}
To implement the \textit{Corruption} stage of our ODPure, we employ the 15 diverse image corruption types established in the robustness benchmark~\cite{Liu2023Detecting}. These corruptions are categorized into four distinct groups to ensure a comprehensive evaluation of trigger stability under various degradations:
\begin{itemize}
    \item \textbf{Noise:} Gaussian Noise, Shot Noise, and Impulse Noise;
    \item \textbf{Blur:} Defocus Blur, Glass Blur, Motion Blur, and Zoom Blur;
    \item \textbf{Weather:} Snow, Frost, Fog, and Brightness;
    \item \textbf{Digital:} Contrast, Elastic Transform, Pixelate, and JPEG Compression.
\end{itemize}
As specified in the main text, each corruption type is applied at three discrete severity levels (1, 2, and 3), resulting in a total of $15 \times 3 = 45$ corrupted variants for each input image. This diverse portfolio allows us to exploit the \textit{variable vulnerability} of triggers across different frequency and spatial domains.

For the ablation studies in Section~\ref{sec:ablation}, unless otherwise stated, experiments are performed on the YOLO detector subjected to the ODA with a standard chessboard trigger of $29\times29$ pixels. The specific baseline variants are configured as follows: the \textit{Single Corruption} variant applies only Defocus Blur at severity level 2; the \textit{w/o Reconstruction} variant omits the reconstruction module; and the \textit{Random Selection} variant replaces the consensus mechanism with random sampling from the purified outputs.

\section{Detailed Results for Adaptive Attacks}\label{secB}

\subsection{Adaptive Attack with Solid Color Trigger}\label{B1}

As hypothesized in the main manuscript, the solid white trigger poses a unique challenge to our defense, primarily targeting the \textit{Corruption} phase. Due to their low-frequency nature and lack of texture, these patterns are less susceptible to the stochastic noise injected during our corruption process. Consequently, the diffusion model may inadvertently interpret the residual trigger patterns as structural content rather than noise, leading to their partial preservation. The following results quantify ODPure's robustness in these adversarial scenarios.

\noindent \textbf{Object Misclassification Attack.}
The results for OMA with the solid-white trigger are presented in Table~\ref{tab:combined_white_trigger}. This adaptive attack proves resilient; for instance, on COCO with the YOLO model, the ASR remains at 62.3\% post-defense. However, focusing solely on ASR obscures the full picture. The attack initially exerts a catastrophic impact on model utility, crippling the mAP to a near-zero \textbf{4.5\%}. Remarkably, while ODPure struggles to fully eliminate the trigger, it successfully salvages the model's utility, achieving a dramatic restoration of the mAP to \textbf{44.6\%}. This indicates that our method recovers the semantic features of legitimate objects, effectively neutralizing the attack's denial-of-service capability even when the trigger partially survives.

\begin{table}[htbp]
\centering
\caption{Robustness against adaptive attacks with the Solid-White Patch Trigger.}
\label{tab:combined_white_trigger}
\scriptsize
\setlength{\tabcolsep}{0pt}
\renewcommand{\arraystretch}{1.1}

\begin{tabular*}{\columnwidth}{@{\extracolsep{\fill}} l l 
    c @{\hspace{1.0em}} 
    c @{\hspace{0.7em}} c 
    @{\hspace{0.2em}} 
    c @{\hspace{0.7em}} c 
    @{}}
\toprule
\multirow{2}{*}{\textbf{Attack}} & 
\multirow{2}{*}{\textbf{Dataset (Model)}} & 
\textbf{Clean} & 
\multicolumn{2}{c}{\textbf{Before Defense}} & 
\multicolumn{2}{c}{\textbf{After Defense}} \\
\cmidrule(r){3-3} \cmidrule{4-5} \cmidrule{6-7}
& & mAP & mAP & ASR & mAP & ASR \\
\midrule
\multirow{4}{*}{OMA} 
& VOC (YOLO)   & 75.5\% & 46.5\% & 82.0\% & 67.9\% & \textbf{52.9\%} \\
& VOC (F-RCNN) & 79.3\% & 48.7\% & 95.2\% & 74.6\% & \textbf{72.6\%} \\
& COCO (YOLO)  & 53.4\% & 4.5\%  & 95.0\% & 44.6\% & \textbf{62.3\%} \\
& COCO (F-RCNN)& 50.3\% & 9.5\%  & 91.1\% & 40.9\% & \textbf{70.9\%} \\
\midrule
\multirow{4}{*}{ODA} 
& VOC (YOLO)   & 71.8\% & 71.3\% & 71.1\% & \textbf{75.6\%} & \textbf{44.0\%} \\
& VOC (F-RCNN) & 80.5\% & 78.0\% & 49.9\% & 80.0\%          & \textbf{24.4\%} \\
& COCO (YOLO)  & 54.1\% & 53.2\% & 96.1\% & 53.1\%          & \textbf{56.4\%} \\
& COCO (F-RCNN)& 50.6\% & 50.1\% & 86.6\% & 50.3\%          & \textbf{38.5\%} \\
\midrule
\multirow{4}{*}{OGA} 
& VOC (YOLO)   & 80.2\% & 77.6\% & 73.7\% & \textbf{83.2\%} & \textbf{37.0\%} \\
& VOC (F-RCNN) & 84.8\% & 82.1\% & 92.7\% & 82.9\%          & \textbf{52.7\%} \\
& COCO (YOLO)  & 55.3\% & 54.3\% & 88.2\% & 54.3\%          & \textbf{57.0\%} \\
& COCO (F-RCNN)& 50.6\% & 49.1\% & 92.6\% & \textbf{51.0\%} & \textbf{51.2\%} \\
\bottomrule
\end{tabular*}
\end{table}

\noindent \textbf{Object Disappearance Attack.}
Against the ODA variant, ODPure demonstrates consistent efficacy in preserving model utility (Table~\ref{tab:combined_white_trigger}). While the ASR reduction is more modest compared to standard trigger attacks (\textit{e.g.,} reducing to 44.0\% for YOLO on VOC), the mAP is consistently preserved or even enhanced. For example, for Faster R-CNN on VOC, the mAP increases from 78.0\% to \textbf{80.0\%} after defense, reinforcing our observation that the reconstruction process contributes to incidental image enhancement.

\noindent \textbf{Object Generation Attack.}
The results for OGA (Table~\ref{tab:combined_white_trigger}) further confirm this trend. ODPure successfully mitigates the attack, reducing the ASR across all scenarios, for instance, from 92.7\% to \textbf{52.7\%} for Faster R-CNN on VOC. Crucially, it filters out falsely generated objects without compromising legitimate detections. The YOLO model on VOC, for example, sees its mAP boosted from 77.6\% to \textbf{83.2\%} post-defense, surpassing the performance of the victim model.

\subsection{Adaptive Attack with Physical-world Trigger}\label{B2}

This attack strategy targets the \textit{Reconstruction} module, aiming to exploit the generative prior by masquerading the trigger as a benign object (\textit{i.e.,} a Pok\'{e} Ball)~\cite{zhangbadrobot}. However, as noted in the main text, the complex texture and high-frequency details of physical triggers often render them more vulnerable to our \textit{Corruption} phase than simple solid patches, preventing their faithful reconstruction.

\noindent \textbf{Object Misclassification Attack.}
As detailed in Table~\ref{tab:adaptive_physical} of the main manuscript, the physical-world trigger causes devastating damage to the undefended model, with the YOLO mAP on COCO plummeting to a negligible \textbf{1.7\%}. While the ASR reduction is less complete (\textit{i.e.,} \textbf{31.1\%} on COCO), likely because the reconstruction module perceives the surviving trigger features as a coherent object, ODPure's ability to restore utility is remarkable. It boosts the mAP from 1.7\% back to \textbf{51.2\%}, effectively reviving the detector's functionality.

\noindent \textbf{Object Disappearance Attack.}
Against ODA, ODPure further demonstrates robust countermeasure capability (see Table~\ref{tab:adaptive_physical}). On the VOC dataset, ODPure not only significantly reduces the ASR for both YOLO and Faster R-CNN (to \textbf{35.1\%} and \textbf{17.8\%}, respectively) but also restores the mAP to levels exceeding the undefended state. For the YOLO model, the mAP is recovered from 72.3\% to \textbf{78.3\%}, surpassing its clean performance.

\noindent \textbf{Object Generation Attack.}
Against OGA, ODPure achieves its most significant success (Table~\ref{tab:adaptive_physical}). It achieves a drastic reduction in ASR, for instance, dropping from 99.6\% to just \textbf{6.7\%} for Faster R-CNN on VOC, while maintaining or improving mAP. This suggests that while the physical trigger is difficult to remove when overlapping with objects (OMA), it is easily identified and filtered out by our pipeline when it appears as an isolated hallucination (OGA).

\end{document}